\documentclass[10pt,twocolumn,letterpaper]{article}

\usepackage{cvpr}              % To produce the CAMERA-READY version
\usepackage{subcaption}
\usepackage{graphicx}
\usepackage{amsmath}
\usepackage{amssymb}
\usepackage{booktabs,multirow}
\usepackage[accsupp]{axessibility}  % Improves PDF readability for those with disabilities.
\usepackage{engord}
\usepackage{seqsplit}
\usepackage[dvipsnames]{xcolor}

\definecolor{cvprblue}{rgb}{0.21,0.49,0.74}
\usepackage[pagebackref,breaklinks,colorlinks,citecolor=cvprblue]{hyperref}

\usepackage[capitalize]{cleveref}
\crefname{section}{Sec.}{Secs.}
\Crefname{section}{Section}{Sections}
\Crefname{table}{Table}{Tables}
\crefname{table}{Tab.}{Tabs.}

\def\confName{CVPR}
\def\confYear{2024}

\begin{document}

\title{NTIRE 2024 Quality Assessment of AI-Generated Content Challenge}

% \author{Yixuan~Gao,
%         Xiongkuo~Min,~\IEEEmembership{Member,~IEEE,}
%         Wenhan~Zhu,
%         Xiao-Ping~Zhang$^{*}$,~\IEEEmembership{Fellow,~IEEE,}
%         and~Guangtao~Zhai$^{*}$,~\IEEEmembership{Senior Member,~IEEE}
               
% % \thanks{Yixuan~Gao, Xiongkuo~Min, Wenhan~Zhu, and Guangtao~Zhai are with Institute of Image Communication and Network Engineering, Shanghai Jiao Tong University, China (E-mail: gaoyixuan@sjtu.edu.cn; minxiongkuo@sjtu.edu.cn; zhuwenhan823@sjtu.edu.cn; zhaiguangtao@sjtu.edu.cn).}% <-this % stops a space
% % \thanks{Xiao-Ping~Zhang is with Tsinghua Berkeley Shenzhen Institute, China and the Department of Electrical, Computer $\&$ Biomedical Engineering, Toronto Metropolitan
% % University, ON M5B 2K3, Canada. (E-mail: xzhang@ryerson.ca).}% <-this % stops a space
% % \thanks{These authors contributed equally to this work.}
% % \thanks{$^{*}$Corresponding co-authors: Guangtao~Zhai and Xiao-Ping~Zhang.}

%         }
\author{
 Xiaohong Liu$^{*}$ \and Xiongkuo Min$^{*}$ \and Guangtao Zhai$^{*}$ \and Chunyi Li$^{*}$ \and  Tengchuan Kou$^{*}$ \and Wei Sun$^{*}$ \and Haoning Wu$^{*}$ \and    Yixuan Gao$^{*}$ \and  Yuqin Cao$^{*}$ \and Zicheng Zhang$^{*}$ \and Xiele Wu$^{*}$ \and  Radu Timofte
 \thanks{The organizers of the NTIRE 2024 Quality Assessment of AI-Generated Content Challenge.\\
 The NTIRE 2024 website: \url{https://cvlai.net/ntire/2024/}.}
\and Fei Peng \and Huiyuan Fu \and Anlong Ming\and Chuanming Wang \and Huadong Ma\and Shuai He \and  Zifei Dou  \and Shu Chen
\and Huacong Zhang\and Haiyi Xie \and Chengwei Wang \and Baoying Chen \and Jishen Zeng \and Jianquan Yang
\and Weigang Wang \and Xi Fang \and Xiaoxin Lv\and Jun Yan
\and Tianwu Zhi \and Yabin Zhang \and Yaohui Li \and Yang Li \and Jingwen Xu \and Jianzhao Liu \and Yiting Liao \and Junlin Li \and Zihao Yu 
\and Fengbin Guan 
\and Yiting Lu
\and Xin Li  
\and Hossein Motamednia\and S. Farhad Hosseini-Benvidi\and Ahmad Mahmoudi-Aznaveh \and Azadeh Mansouri 
\and Ganzorig Gankhuyag \and Kihwan Yoon 
\and Yifang Xu \and Haotian Fan \and Fangyuan Kong
\and Shiling Zhao \and Weifeng Dong\and Haibing Yin
\and Li Zhu \and Zhiling Wang \and Bingchen Huang \and Avinab Saha \and Sandeep Mishra \and Shashank Gupta \and Rajesh Sureddi \and Oindrila
Saha
\and Luigi Celona \and Simone Bianco \and Paolo Napoletano \and Raimondo Schettini 
\and Junfeng Yang \and Jing Fu \and Wei Zhang \and Wenzhi Cao \and Limei Liu \and Han Peng
\and Weijun Yuan \and Zhan Li \and Yihang Cheng \and Yifan Deng 
\and Haohui Li \and Bowen Qu \and Yao Li \and Shuqing Luo \and Shunzhou Wang \and Wei Gao 
\and Zihao Lu \and Marcos V. Conde \and Radu Timofte 
\and Xinrui Wang \and Zhibo Chen \and Ruling Liao \and Yan Ye 
\and Qiulin Wang \and Bing Li \and Zhaokun Zhou \and Miao Geng \and Rui Chen \and Xin Tao 
\and Xiaoyu Liang \and Shangkun Sun 
\and Xingyuan Ma 
\and Jiaze Li \and Mengduo Yang \and Haoran Xu  \and Jie Zhou \and Shiding Zhu \and Bohan Yu 
\and Pengfei Chen \and Xinrui Xu \and Jiabin Shen  \and Zhichao Duan 
\and Erfan Asadi 
\and Jiahe Liu  \and Qi Yan \and Youran Qu \and Xiaohui Zeng \and Lele Wang \and Renjie Liao
%\thanks{The valid participating teams of the NTIRE 2024 Quality Assessment of AI-Generated Content Challenge.\\
%The NTIRE 2024 website: \url{https://cvlai.net/ntire/2024/}}
% \thanks{Pengxiang Xiao is also Vrobotit Lab, Beijing University of Posts and Telecommunications and the work is primarily done during an internship at Alibaba Group.}
}
\maketitle

%%%%%%%%% ABSTRACT
\begin{abstract}
This paper reports on the NTIRE 2024 Quality Assessment of AI-Generated Content Challenge, which will be held in conjunction with the New Trends in Image Restoration and Enhancement Workshop (NTIRE) at CVPR 2024. This challenge is to address a major challenge in the field of image and video processing, namely, Image Quality Assessment (IQA) and Video Quality Assessment (VQA) for AI-Generated Content (AIGC). The challenge is divided into the image track and the video track.
The image track uses the AIGIQA-20K, which contains 20,000 AI-Generated Images (AIGIs) generated by 15 popular generative models. The image track has a total of 318 registered participants. A total of 1,646 submissions are received in the development phase, and 221 submissions are received in the test phase. Finally, 16 participating teams submitted their models and fact sheets.

The video track uses the T2VQA-DB, which contains 10,000 AI-Generated Videos (AIGVs) generated by 9 popular Text-to-Video (T2V) models. A total of 196 participants have registered in the video track. A total of 991 submissions are received in the development phase, and 185 submissions are received in the test phase. Finally, 12 participating teams submitted their models and fact sheets. Some methods have achieved better results than baseline methods, and the winning methods in both tracks have demonstrated superior prediction performance on AIGC.

\end{abstract}

%%%%%%%%% BODY TEXT
\section{Introduction}
\label{sec:intro}

With the fast development of generative models, AI-Generated Content (AIGC) has become popular in daily lives. Among them, AI-Generated Images (AIGIs) and AI-Generated Videos (AIGVs) are two of the most common media. However, the quality of AIGIs and AIGVs can be varied due to the differences in performance of various models. Therefore, it is significant to propose efficient Image Quality Assessment (IQA) and Vmage Quality Assessment (VQA) methods to accurately predict the quality of
generated images and videos 

This NTIRE 2024 Quality Assessment of AI-Generated Content Challenge aims to promote the development of the I/VQA methods for generated images and videos to guide the improvement and enhancement of the performance of generative models, thereby improving the quality of experience of AIGC. The challenge is divided into the image track and the video track. In the image track, we use the AIGIQA-20K~\cite{AIGIQA-20K}, which contains 20,000 AIGIs generated by 15 Text-to-Image (T2I) models. 21 subjects are invited to produce accurate Mean Opinion Scores (MOSs). The video track uses the T2VQA-DB~\cite{kou2024subjective}, in which 9 Text-to-Vido (T2V) models are used to generate 10,000 videos. The MOSs are obtained from 27 subjects. 

This is the first time that a quality assessment of AIGC challenge has been held at the NTIRE workshop. The challenge has a total of 514 registered participants, 318 in the image track and 196 in the video track. A total of 2,637 submissions were received in the development phase, while 406 prediction results were submitted during the final testing phase. Finally, 16 valid participating teams in the image track and 12 valid participating teams in the video track submitted their final models and fact sheets. They have provided detailed introductions to their I/VQA methods for AIGIs and AIGVs. We provide the detailed results of the challenge in Section~\ref{Challenge Results} and Section~\ref{Challenge Methods}. We hope that this challenge can promote the development of I/VQA methods for image and video generation.

This challenge is one of the NTIRE 2024 Workshop~\footnote{\url{https://cvlai.net/ntire/2024/}} series of challenges on: dense and non-homogeneous dehazing, night photography rendering, blind compressed image enhancement, shadow removal, efficient super-resolution, image super-resolution ($\times 4$), light field image super-resolution, stereo image super-resolution, HR depth from images of specular and transparent surfaces, bracketing image restoration and enhancement, portrait quality assessment, Restore Any Image Model (RAIM) in the wild, raW image super-resolution, short-form UGC Video quality assessment, low light enhancement, and raw burst alignment and ISP challenge.

 % This challenge is one of the NTIRE 2024 Workshop~\footnote{\url{https://cvlai.net/ntire/2024/}} series of challenges on: night photography rendering~\cite{shutova2023ntire_night}, HR depth from images of specular and transparent surfaces~\cite{zama2023ntire_depth}, image denoising~\cite{li2023ntire_dn50}, video colorization~\cite{kang2023ntire_vc}, shadow removal~\cite{vasluianu2023ntire_isr}, quality assessment of video enhancement~\cite{liu2023ntire}, stereo super-resolution~\cite{wang2023ntire_ssr}, light field image super-resolution~\cite{wang2023ntire_lfsr}, image super-resolution ($\times4$)~\cite{zhang2023ntire}, 360° omnidirectional image and video super-resolution~\cite{cao2023ntire}, lens-to-lens bokeh effect transformation~\cite{conde2023ntire_bokeh}, real-time 4K super-resolution~\cite{conde2023ntire_rtsr}, HR nonhomogenous dehazing~\cite{ancuti2023ntire}, efficient super-resolution~\cite{li2023ntire_esr}.

\section{Related Work}

\subsection{AIGI dataset}

Several AIGI datasets have been proposed in recent years. Benefiting from the successful Stable Diffusion~\cite{rombach2022high}, DiffusionDB~\cite{wang2022diffusiondb} is the first large-scale text-to-image prompt dataset, containing 14 million images generated by Stable Diffusion using prompts and hyperparameters specified by real users. HPS~\cite{wu2023human} collects 98,807 generated images from the Stable Foundation Discord channel, along with 25,205 human choices.  ImageReward~\cite{xu2024imagereward} proposes a dataset containing 137k prompt-image pairs sampled from DiffusionDB. Each pair has 3 MOSs from overall rating, image-text alignment, and fidelity. Pick-A-Pic~\cite{kirstain2024pick} contains over 500,000 examples and 35,000 distinct prompts, Each example contains a prompt, two generated images, and a label for which image is preferred. AGIQA-1K~\cite{zhang2023perceptual}, AGIQA-3K~\cite{li2023agiqa}, and AIGCIQA2023~\cite{wang2023aigciqa2023} contain 1,080, 2,982, and 2,400 images respectively. AGIN~\cite{chen2023exploring} collects 6,049 images and conducts a large-scale subjective study to collect human opinions on the overall naturalness. In this challenge, we use the AIGIQA-20K~\cite{AIGIQA-20K}, including 20,000 images generated by 15 popular T2I models, along with the MOSs collected from 21 subjects.  

\subsection{AIGV dataset}
Compared with AIGI datasets, the number of proposed AIGV datasets is small. Chivileva \etal~\cite{chivileva2023measuring} proposes a dataset with 1,005 videos generated by 5 T2V models. 24 users are involved in the subjective study. EvalCrafter~\cite{liu2023evalcrafter} builds a dataset using 500 prompts and 5 T2V models, resulting in 2,500 videos in total. However, only 3 users are involved in the subjective study. Similarly, FETV~\cite{liu2024fetv} uses 619 prompts, 4 T2V models, and 3 users for annotation as well. VBench~\cite{huang2024vbench} has a larger scale with in total of $\sim$1,7k prompts and 4 T2V models. In the video track, we use the T2VQA-DB~\cite{kou2024subjective}. The dataset has 10,000 videos generated by 9 different T2V models. 27 subjects are invited to collect the MOSs. 
 
%This dataset is used to test the performance of methods proposed by different participating teams in this challenge.

\subsection{IQA model}

Traditional IQA models focus on distortions like noises, blurriness, semantic contents, etc. DBCNN~\cite{zhang2020blind} handles both synthetic and authentic distortions by training two CNN networks. StairIQA~\cite{sun2023blind} proposes a staircase structure to hierarchically integrate the information from low-level to high-level. LIQE~\cite{zhang2023liqe} proposes a general and automated multitask learning scheme to exploit auxiliary knowledge from IQA, scene classification, and distortion type identification. In the meantime, several IQA models designed for AIGIs have been proposed. HPS~\cite{wu2023human} and PickScore~\cite{kirstain2024pick} are CLIP-based~\cite{radford2021clip} models to imitate human preference on generated images. ImageReward~\cite{xu2024imagereward} uses a BLIP-based~\cite{li2022blip} architecture to predict the image quality. 

In recent years, researchers have been paying attention to using the ability of Large Multi-modality Models (LMMs) to solve IQA tasks. Q-bench~\cite{wu2023qbench} first investigates the performance of LMMs in evaluating visual quality. ~\cite{wu2023qinstruct,zhang2023qboost, wu2023qalign, wu2024towards} further introduce the training procedure to utilize LMMs for IQA tasks. Q-Refine~\cite{li2024qrefine} is a quality-awarded refiner to guide the refining process in T2I models. The development of IQA models not only provides more accurate predictions on AIGIs quality but also benefits the development of image generation models.

\subsection{VQA model}
%The traditional VQA methods are handcrafted feature-based models. This kind of methods first calculate the quality of each frame of a video by extracting quality features, and then obtain the video quality score \cite{min2018blind,zhai2020perceptual,gao2022image}. For example, V-BLIINDS \cite{saad2014blind} is a spatio-temporal natural scene statistics (NSS) model, which can quantify motion coherency in video scenes. TLVQM \cite{korhonen2019two} is based on the idea of calculating features at two levels, that is, first calculating the low complexity features of the entire sequence, and then extracting high complexity features from subsets of representative video frames. VIDEVAL \cite{tu2021ugc} calculates video quality by extracting abundant spatio-temporal features such as motion, jerkiness, blurriness, noise, blockiness, color, and so on. RAPIQUE \cite{tu2021rapique} combines the advantages of both quality-aware scene statistics features and semantics-aware deep convolutional features to calculate video quality.
% FastVQA \cite{wu2022fast} 

The traditional VQA models are usually designed for user-generated videos or a certain attribute of videos.~\cite{gao2023vdpve, dong2023light, kou2023stablevqa,zhang2023advancing,zhang2024reduced,liu2023ntire}. For example, %VSFA \cite{li2019quality} first extracts semantic features from a pre-trained convolutional neural network, and then uses a gated recursive unit network to extract the temporal relationship between semantic features of video frames to predict video quality. BVQA \cite{liu2021spatiotemporal} uses a feature encoder to directly extract spatio-temporal representations from videos to predict video quality.
SimpleVQA \cite{sun2022deep} trains an end-to-end spatial feature extraction network to directly learn quality-aware spatial features from video frames, and extracts motion features to measure temporally related distortions at the same time to predict video quality. FAST-VQA~\cite{wu2022fast} proposes the ``fragments'' sampling strategies and the Fragment Attention Network (FANet) to accommodate fragments as inputs. DOVER~\cite{wu2023dover} evaluates the quality of videos from the technical and aesthetic perspectives respectively. Q-Align~\cite{wu2023qalign} can also address the VQA task by relying on the ability of multi-modal large models. 

There are several works targeting the VQA tasks of AIGVs. VBench~\cite{huang2024vbench} and EvalCrafter~\cite{liu2023evalcrafter} build benchmarks for AIGVs by designing multi-dimensional metrics. MaxVQA~\cite{wu2023maxvqa} and FETV~\cite{liu2024fetv} propose separate metrics for the assessment of video-text alignment and video fidelity, while T2VQA~\cite{kou2024subjective} handles the features from the two dimensions as a whole. We believe the development of the VQA model for AIGV will certainly benefit the generation of high-quality videos.

\section{NTIRE 2024 Quality Assessment of AI-Generated Content Challenge}
We organize the NTIRE 2024 Quality Assessment of AI-Generated Content Challenge in order to promote the development of objective I/VQA methods for AIGIs and AIGVs. The main goal of the challenge is to predict the perceptual quality of the generated images and videos. Details about the challenge are as follows:

\subsection{Overview}
The challenge has two tracks, \ie image track and video track. The task is to predict the perceptual quality of a generated image/video based on a set of prior examples of images/videos and their perceptual quality labels. The challenge uses the AIGIQA-20K~\cite{AIGIQA-20K} and the T2VQA-DB~\cite{kou2024subjective} dataset and splits them into the training, validation, and testing sets. As the final result, the participants in the challenge are asked to submit predicted scores for the given testing set.

\subsection{Datasets}

\begin{table*}[!t]
    \centering
    \caption{Quantitative results for the NTIRE 2024 Quality Assessment of AI-Generated Content Challenge: Track 1 Image.}
    \resizebox{0.8\textwidth}{!}{
    \begin{tabular}{c|c|c|ccc}
    \toprule
    Rank & Team & Leader & Main Score & SRCC & PLCC \\
    \midrule
    1 & pengfei & Fei Peng & 0.9175 & 0.9076 & 0.9274\\
    2 & MediaSecurity\_SYSU\&Alibaba & Huacong Zhang & 0.9169 & 0.9076 & 0.9262 \\
    3 & geniuswwg & Weigang Wang & 0.9157 & 0.9051 & 0.9264 \\
    4 & Yag & Zihao Yu & 0.9138 & 0.9009 & 0.9268 \\
    5 & QA-FTE & Tianwu Zhi & 0.9091 & 0.8982 & 0.9201 \\
    6 & HUTB-IQALab & Junfeng Yang & 0.9087 & 0.8957 & 0.9218 \\
    7 & IQ Analyzers & Avinab Saha & 0.9065 & 0.8912 & 0.9217 \\
    8 & PKUMMCAL & Haohui Li & 0.9044 & 0.8933 & 0.9155 \\
    9 & BDVQAGroup & Yifang Xu & 0.9023 & 0.8926 & 0.9119 \\
    10 & JNU\_620 & Weijun Yuan & 0.8835 & 0.8746 & 0.8923 \\ 
    11 & MT-AIGCQA & Li Zhu & 0.8736 & 0.8589 & 0.8883 \\
    12 & IVL & Luigi Celona & 0.8715 & 0.8486 & 0.8944 \\
    13 & CVLab & Zihao Lu & 0.8657 & 0.8522 & 0.8792 \\
    14 & z6 & Ganzorig Gankhuyag & 0.8628 & 0.8472 & 0.8785 \\
    15 & Oblivion & Shiling Zhao & 0.8613 & 0.8751 & 0.8476 \\
    16 & IVP-Lab & Hossein Motamednia & 0.8595 & 0.8429 & 0.8762 \\
    \midrule
    \multirow{3}{*}{Baseline} & \multicolumn{2}{c|}{StairIQA~\cite{sun2023blind}} & 0.637 & 0.6179 & 0.6561 \\
    ~ & \multicolumn{2}{c|}{DBCNN~\cite{zhang2020blind}} & 0.8228 & 0.7914 & 0.8542\\
    ~ & \multicolumn{2}{c|}{LIQE~\cite{zhang2023liqe}} & 0.8543 & 0.8652 & 0.8433 \\
    \bottomrule
    \end{tabular}
    }
    \label{tab:image results}
\end{table*}

\begin{table*}[!t]
    \centering
    \caption{Quantitative results for the NTIRE 2024 Quality Assessment of AI-Generated Content Challenge: Track 2 Video.}
    \resizebox{0.8\textwidth}{!}{
    \begin{tabular}{c|c|c|ccc}
    \toprule
    Rank & Team & Leader & Main Score & SRCC & PLCC \\
    \midrule
    1 & IMCL-DAMO & Yiting Lu & 0.8385 & 0.8322 & 0.8448\\
    2 & Kwai-kaa & Qiulin Wang & 0.824 & 0.8154 & 0.8326 \\
    3 & SQL & Wei Gao & 0.8232 & 0.8148 & 0.8316 \\
    4 & musicbeer & Xiaoxin Lv & 0.8231 & 0.8144 & 0.8318 \\
    5 & finnbingo & Xingyuan Ma & 0.8211 & 0.8131 & 0.829 \\
    6 & PromptSync & Jiaze Li & 0.8178 & 0.8102 & 0.8254 \\
    7 & QA-FTE & Tianwu Zhi & 0.8128 & 0.805 & 0.8207 \\
    8 & MediaSecurity\_SYSU\&Alibaba & Baoying Chen & 0.8124 & 0.8021 & 0.8226 \\
    9 & IPPL-VQA & Pengfei Chen & 0.8003 & 0.7939 & 0.8066 \\
    10 & IVP-Lab & Hossein Motamednia & 0.7944 & 0.7852 & 0.8035 \\ 
    11 & Oblivion & Weifeng Dong & 0.7869 & 0.7773 & 0.7965 \\
    %12 & CUC-IMC & Zelu Qi & 0.7802 & 0.7695 & 0.791 \\
    12 & UBC DSL Team & Jiahe Liu & 0.7531 & 0.7431 & 0.7632 \\
    \midrule
    \multirow{3}{*}{Baseline} & \multicolumn{2}{c|}{SimpleVQA~\cite{sun2022deep}} & 0.6602 & 0.6489 & 0.6714 \\
    ~ & \multicolumn{2}{c|}{FAST-VQA~\cite{wu2022fast}} & 0.7197 & 0.7156 & 0.7238\\
    ~ & \multicolumn{2}{c|}{DOVER~\cite{wu2023dover}} & 0.7698 & 0.7616 & 0.7779 \\
    \bottomrule
    \end{tabular}
    }
    \label{tab:video results}
\end{table*}
% We construct a new dataset called VQA Dataset for Perceptual Video Enhancement (VDPVE).

In the image track, we use the AIGIQA-20K~\cite{AIGIQA-20K} for training, validating, and testing. The dataset contains 20,000 images generated by 15 T2I models, which are: DALLE 2~\cite{gen:dalle}, DALLE 3~\cite{gen:dalle}, Dreamlike~\cite{gen:dream}, IF~\cite{gen:IF}, LCM Pixart~\cite{gen:lcm}, LCM SD1.5~\cite{gen:lcm}, LCM SDXL~\cite{gen:lcm}, Midjourney v5.2~\cite{gen:MJ}, Pixart $\alpha$~\cite{gen:pixart}, Playground v2~\cite{gen:Playground}, SD1.4~\cite{gen:sd}, SD1.5~\cite{gen:sd}, SDXL~\cite{gen:xl}, SDXL Turbo~\cite{gen:turbo}, and SSD1B~\cite{gen:ssd-1b}. Concretely,  20,000 prompts are selected from DiffusionDB~\cite{wang2022diffusiondb}. For Dreamlike, Pixart $\alpha$, Playground v2, SD1.4, SD1.5, SDXL, and SSD1B~\cite{gen:dream,gen:pixart,gen:Playground,gen:sd,gen:ssd-1b,gen:xl}, each model generates 2,000 images for their strong generalize ability. LCM Pixart, LCM SD1.5, LCM SDXL, and SDXL Turbo~\cite{gen:lcm, gen:turbo} are assigned 1,000 images each, and DALLE2, DALLE3, IF, Midjourney v5.2~\cite{gen:dalle, gen:IF, gen:MJ} are 500 images each. 

In the video track, we use the T2VQA-DB~\cite{kou2024subjective}. The dataset contains 10,000 generated videos from: Text2Video-Zero~\cite{khachatryan2023text2video}, AnimateDiff~\cite{guo2023animatediff}, Tune-a-video~\cite{wu2023tune}, VidRD~\cite{gu2023reuse}, VideoFusion~\cite{luo2023videofusion}, ModelScope~\cite{wang2023modelscope}, LVDM~\cite{he2022latent}, Show-1~\cite{zhang2023show}, and LaVie~\cite{wang2023lavie}. 1,000 prompts are selected from WebVid-10M~\cite{Bain21}, a large-scale text-video dataset. Each model generates one video for each prompt. For Tune-a-video~\cite{wu2023tune}, two different pre-trained weights are used. The video resolution is unified to $512 \times 512$, and 
the video length is 4s. 

% The VDPVE has 1211 videos with different enhancements, which can be divided into three sub-datasets: the first sub-dataset has 600 videos with color, brightness, and contrast enhancements; the second sub-dataset has 310 videos with deblurring; and the third sub-dataset has 301 deshaked videos. The resolution of all videos in the VDPVE is $1280\times720$. The video length is 8s or 10s.

% In the first sub-dataset, eight enhancement methods are utilized to enhance the color, brightness, and contrast of 79 videos: ACE \cite{getreuer2012automatic}, AGCCPF \cite{gupta2016minimum}, BPHEME \cite{wang2005brightness}, MBLLEN \cite{lv2018mbllen}, SGZSL \cite{zheng2022semantic}, DCC-Net \cite{zhang2022deep}, and two commercial software: CapCut and Adobe Premiere Pro. In the second sub-dataset, we utilize five enhancement methods to deblur the $62$ blurred videos, including ESTRNN \cite{zhong2020efficient}, DeblurGANv2 \cite{kupyn2019deblurgan}, FGST \cite{lin2022flow}, BasicVSR++ \cite{chan2022basicvsr++}, and Adobe Premiere Pro. In the third sub-dataset, seven enhancement methods are utilized to stabilize $43$ videos, including GlobalFlowNet \cite{james2023globalflownet}, DIFRINT \cite{choi2020deep}, PWStableNet \cite{zhao2020pwstablenet}, Yu \cite{yu2020learning}, CapCut (most stable mode), CapCut (minimum cropping mode), and Adobe Premiere Pro. 
$21$ subjects are invited to rate the generated images in AIGIQA-20K~\cite{AIGIQA-20K}, and $27$ subjects for the videos in T2VQA-DB~\cite{kou2024subjective}. After normalizing and averaging the subjective opinion scores, the mean opinion score (MOS) of each image/video can be obtained. Furthermore, we randomly split the AIGIQA-20K into a training set, a validation set, and a testing set according to the ratio of $7:1:2$. The same split is conducted to the T2VQA-DB. The numbers of generated images in the training set, validation set, and testing set are $14,000$, $2,000$, and $4,000$, respectively. For the video track, the numbers are $7,000$, $1,000$, and $2,000$.

\subsection{Evaluation protocol}

In both tracks, the main scores are utilized to determine the rankings of participating teams. We ignore the sign and calculate the average of Spearman Rank-order Correlation Coefficient (SRCC) and Person Linear Correlation Coefficient (PLCC) as the main score:
\begin{equation}
    \mathrm{Main\;Score} = (|\mathrm{SRCC}| + |\mathrm{PLCC}|)/2.
\end{equation}
SRCC measures the prediction monotonicity, while PLCC measures the prediction accuracy. Better VQA methods should have larger SRCC and PLCC values. Before calculating PLCC index, we perform the third-order polynomial nonlinear regression. By combining SRCC and PLCC, the main scores can comprehensively measure the performance of participating methods.
% as suggested in the previous works [48, 26]

\begin{figure*}[ht]
\centering
\begin{subfigure}{0.23\textwidth}
    \includegraphics[width=\textwidth]{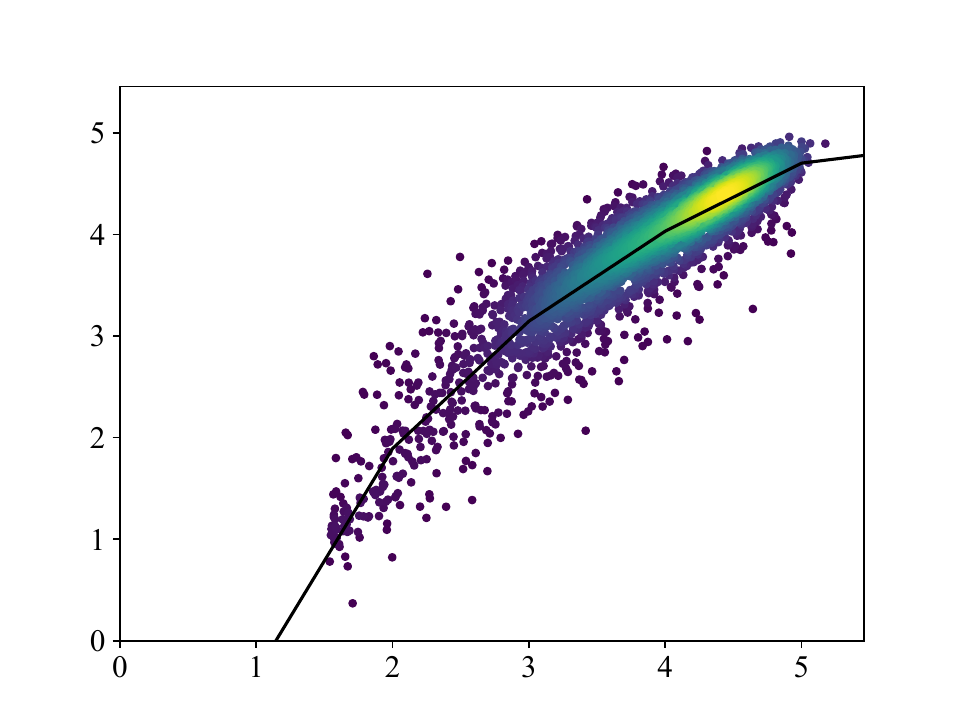}
    \caption{pengfei}
    \label{fig:first}
\end{subfigure}
\hfill
\begin{subfigure}{0.23\textwidth}
    \includegraphics[width=\textwidth]{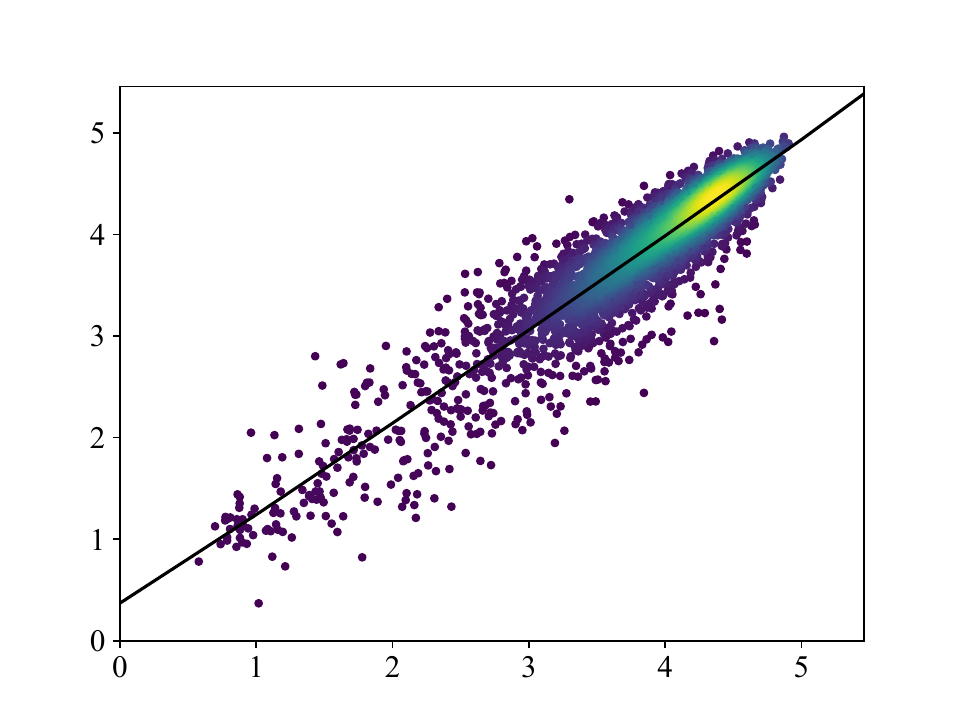}
    \caption{MediaSecurity\_SYSU\&Alibaba}
    \label{fig:first}
\end{subfigure}
\hfill
\begin{subfigure}{0.23\textwidth}
    \includegraphics[width=\textwidth]{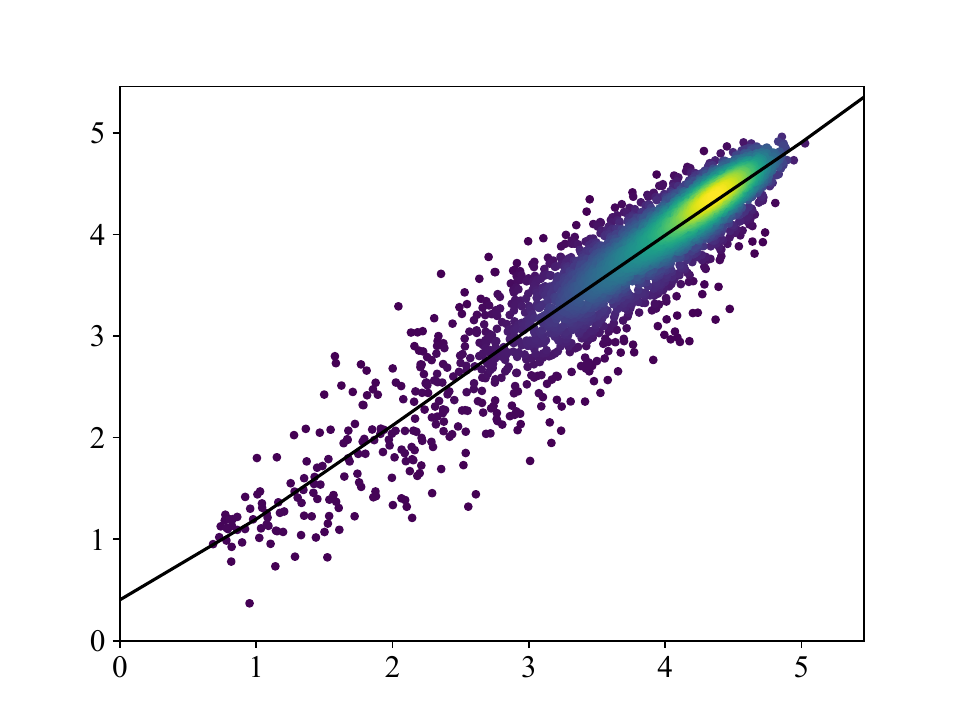}
    \caption{geniuswwg}
    \label{fig:first}
\end{subfigure}
\hfill
\begin{subfigure}{0.23\textwidth}
    \includegraphics[width=\textwidth]{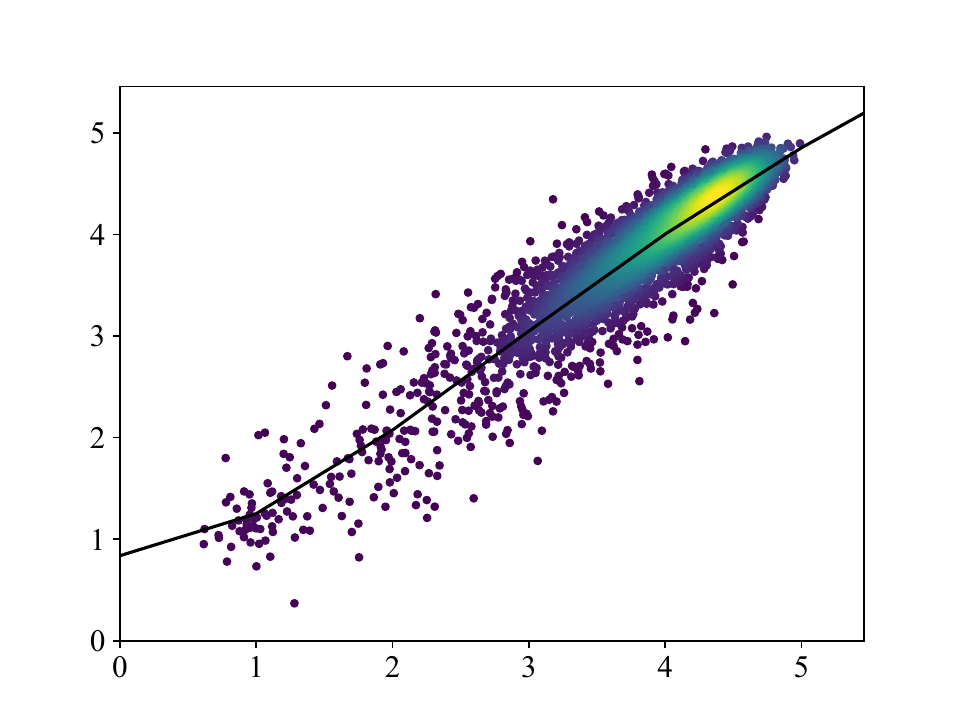}
    \caption{Yag}
    \label{fig:first}
\end{subfigure}
\\
\begin{subfigure}{0.23\textwidth}
    \includegraphics[width=\textwidth]{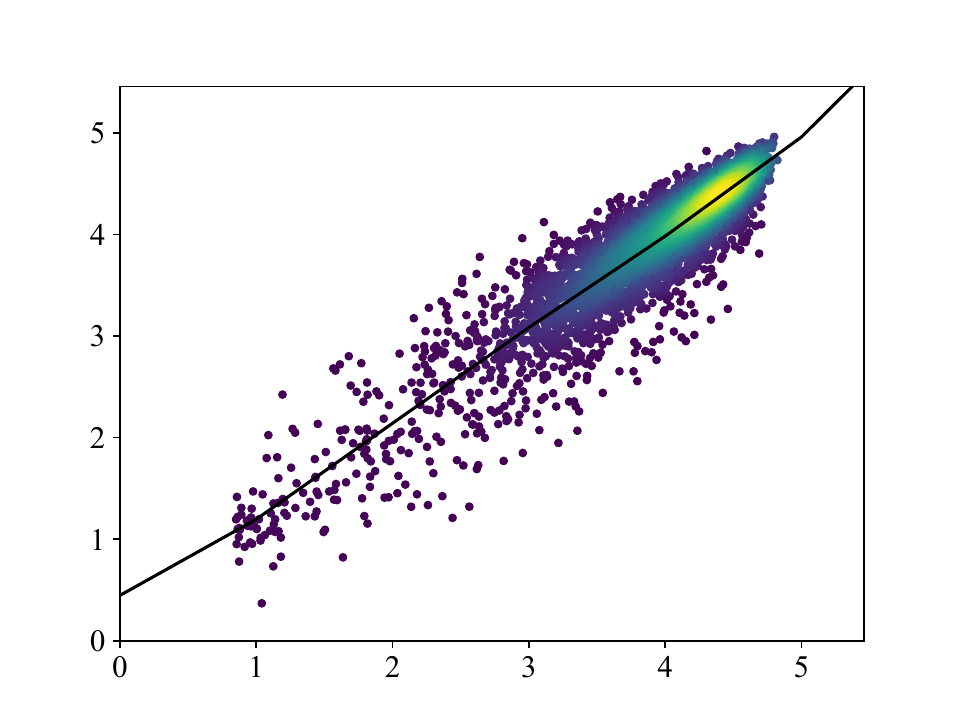}
    \caption{QA-FTE}
    \label{fig:first}
\end{subfigure}
\hfill
\begin{subfigure}{0.23\textwidth}
    \includegraphics[width=\textwidth]{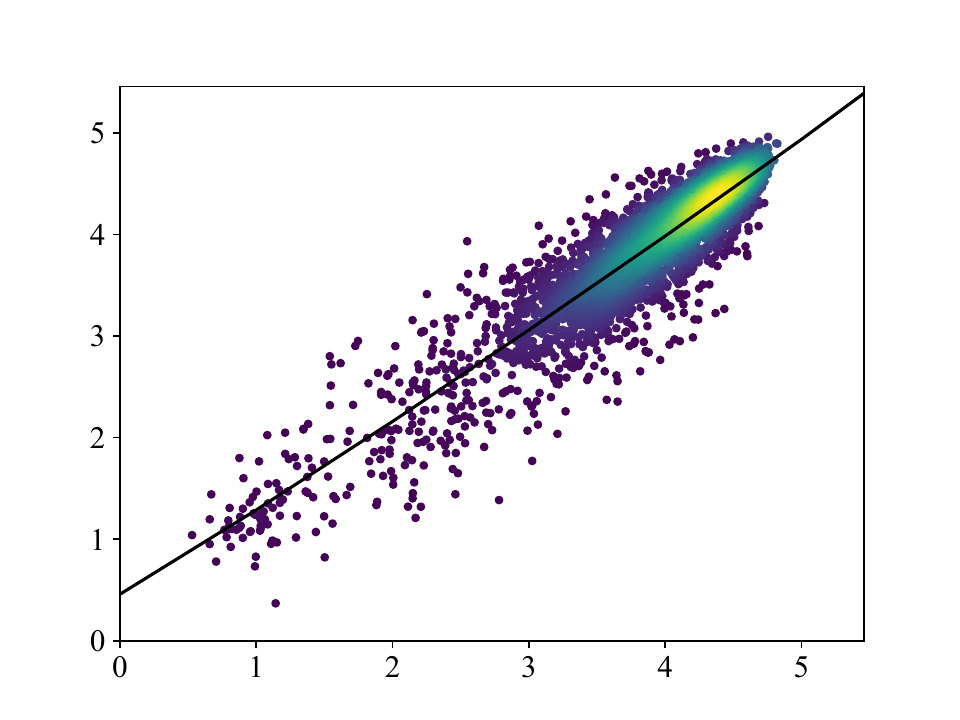}
    \caption{HUTB-IQALab}
    \label{fig:first}
\end{subfigure}
\hfill
\begin{subfigure}{0.23\textwidth}
    \includegraphics[width=\textwidth]{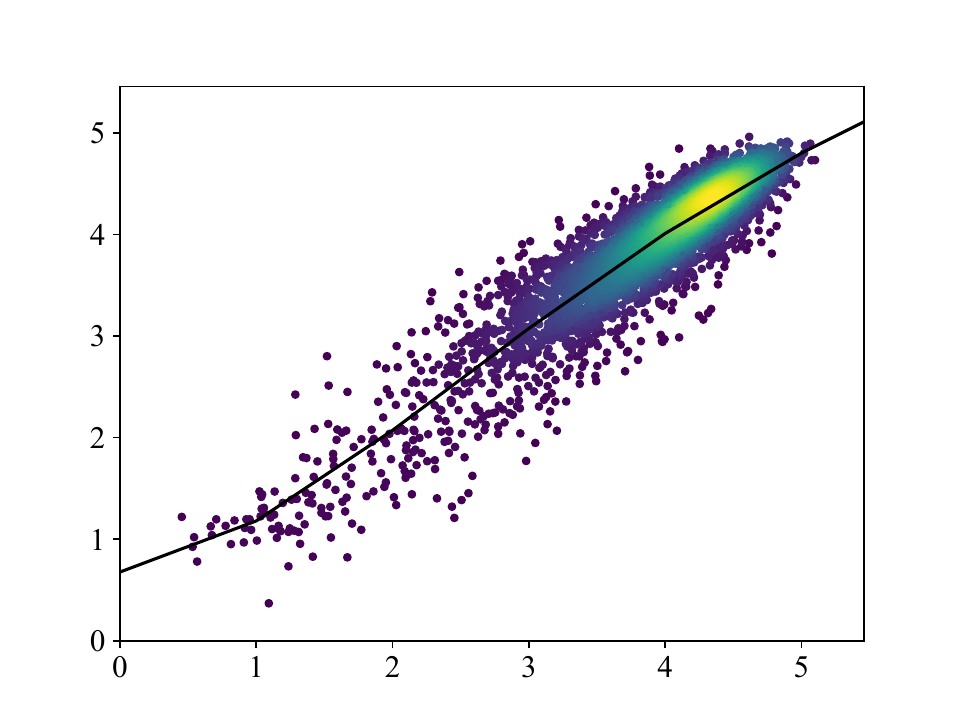}
    \caption{IQAnalyzers}
    \label{fig:first}
\end{subfigure}
\hfill
\begin{subfigure}{0.23\textwidth}
    \includegraphics[width=\textwidth]{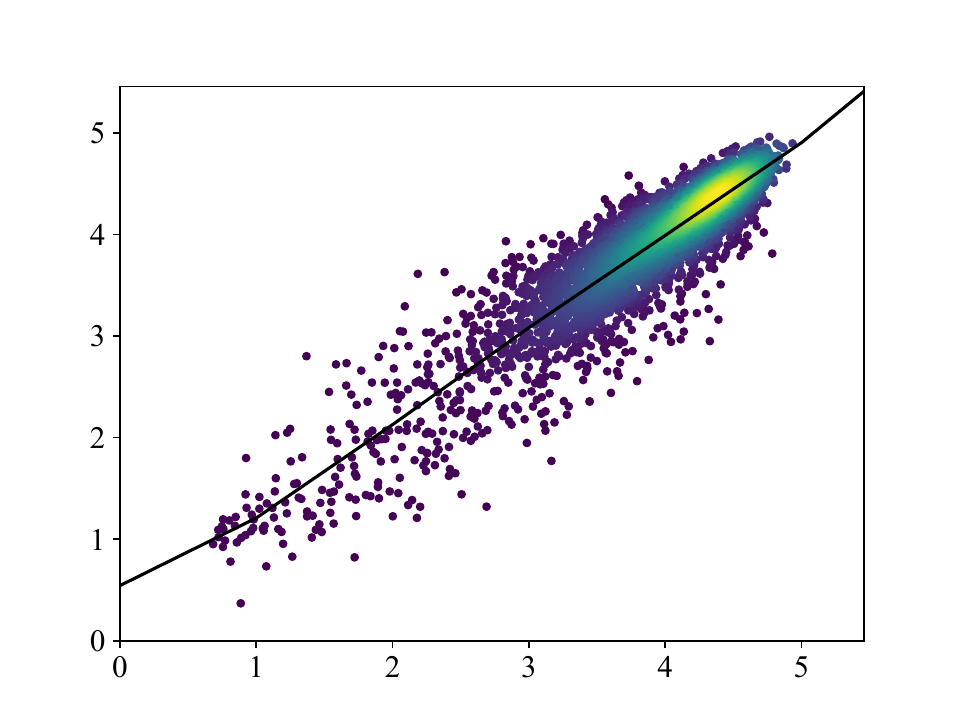}
    \caption{PKUMMCA}
    \label{fig:first}
\end{subfigure}
\\
\begin{subfigure}{0.23\textwidth}
    \includegraphics[width=\textwidth]{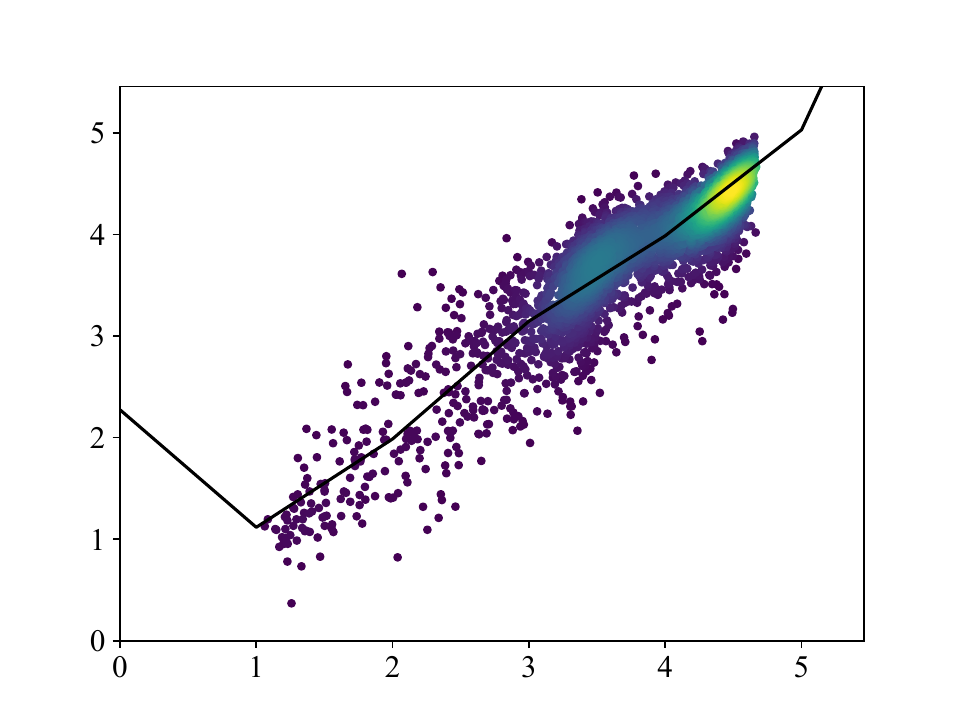}
    \caption{BDVQAGroup}
    \label{fig:first}
\end{subfigure}
\hfill
\begin{subfigure}{0.23\textwidth}
    \includegraphics[width=\textwidth]{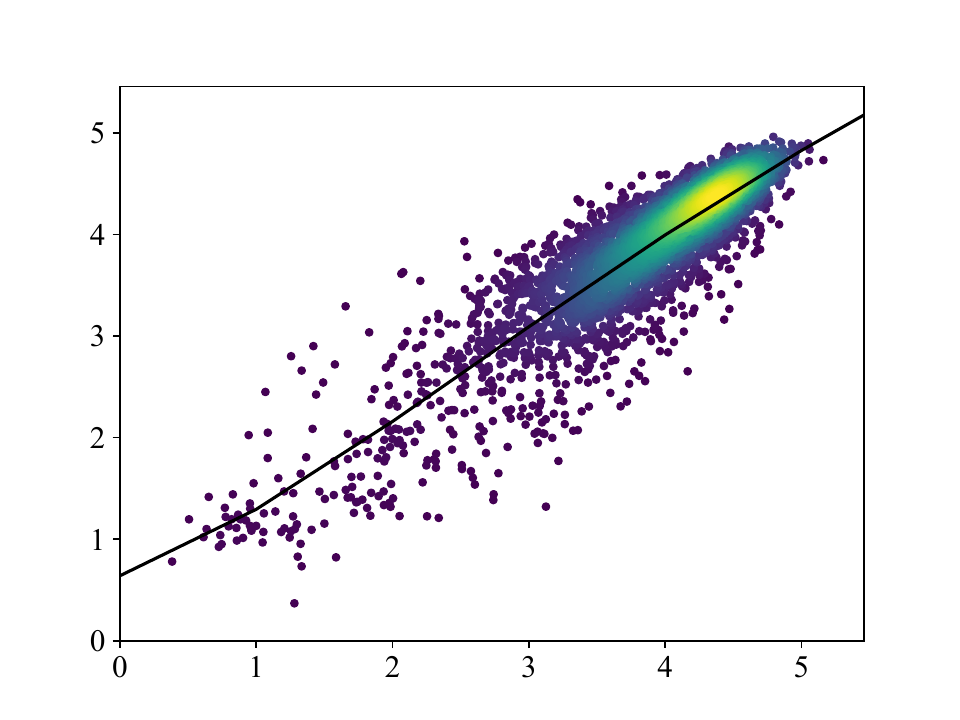}
    \caption{JNU\_620}
    \label{fig:first}
\end{subfigure}
\hfill
\begin{subfigure}{0.23\textwidth}
    \includegraphics[width=\textwidth]{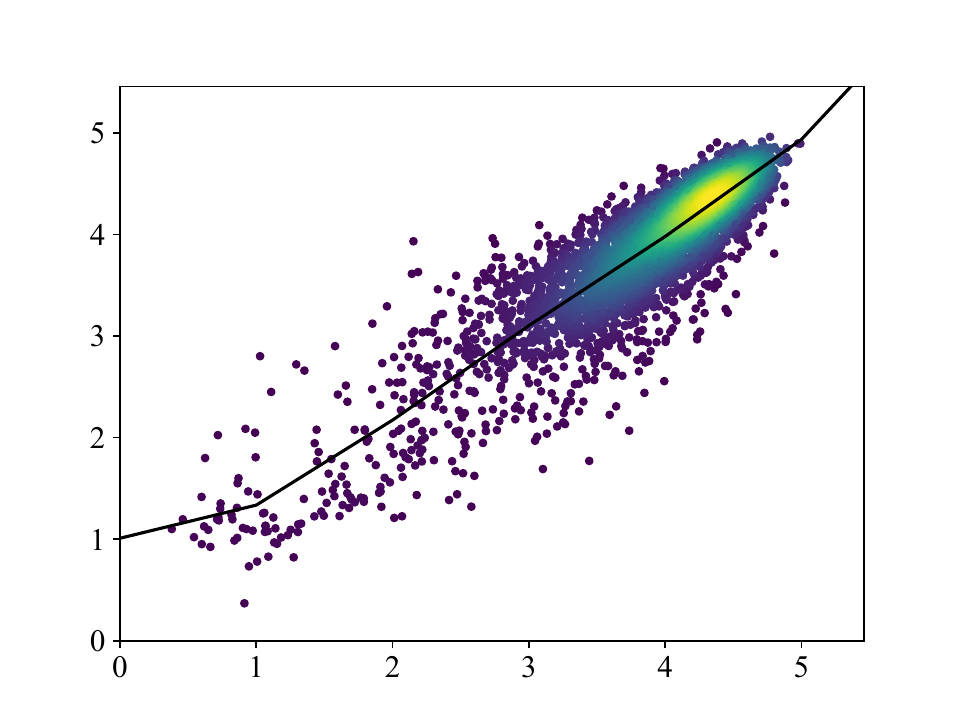}
    \caption{MT-AIGCQA}
    \label{fig:first}
\end{subfigure}
\hfill
\begin{subfigure}{0.23\textwidth}
    \includegraphics[width=\textwidth]{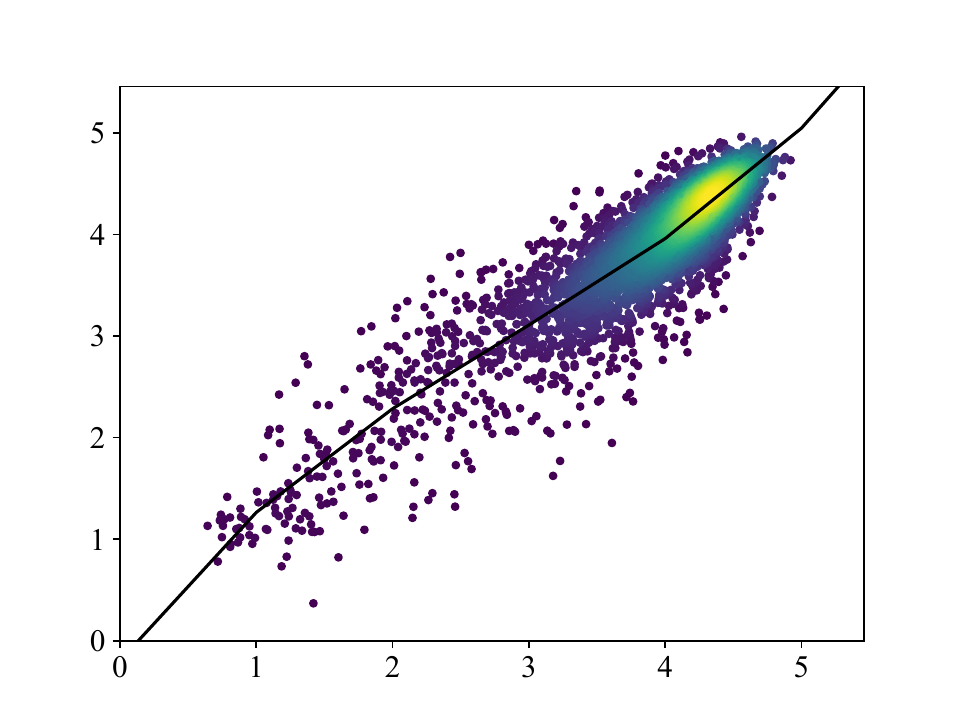}
    \caption{IVL}
    \label{fig:first}
\end{subfigure}
\\
\begin{subfigure}{0.23\textwidth}
    \includegraphics[width=\textwidth]{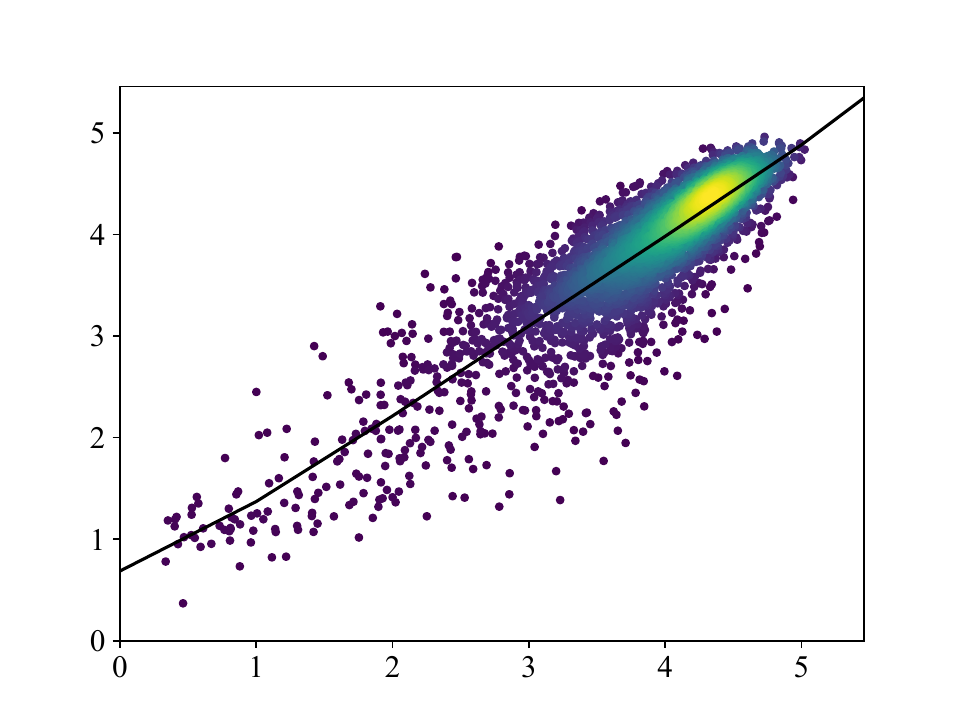}
    \caption{CVLab}
    \label{fig:first}
\end{subfigure}
\hfill
\begin{subfigure}{0.23\textwidth}
    \includegraphics[width=\textwidth]{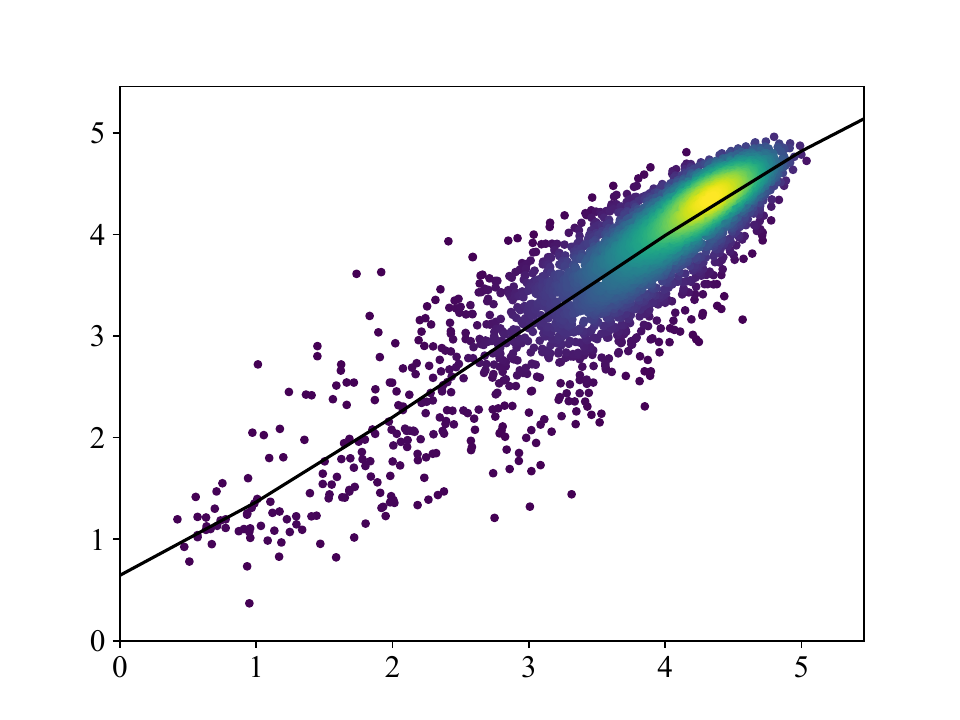}
    \caption{z6}
    \label{fig:first}
\end{subfigure}
\hfill
\begin{subfigure}{0.23\textwidth}
    \includegraphics[width=\textwidth]{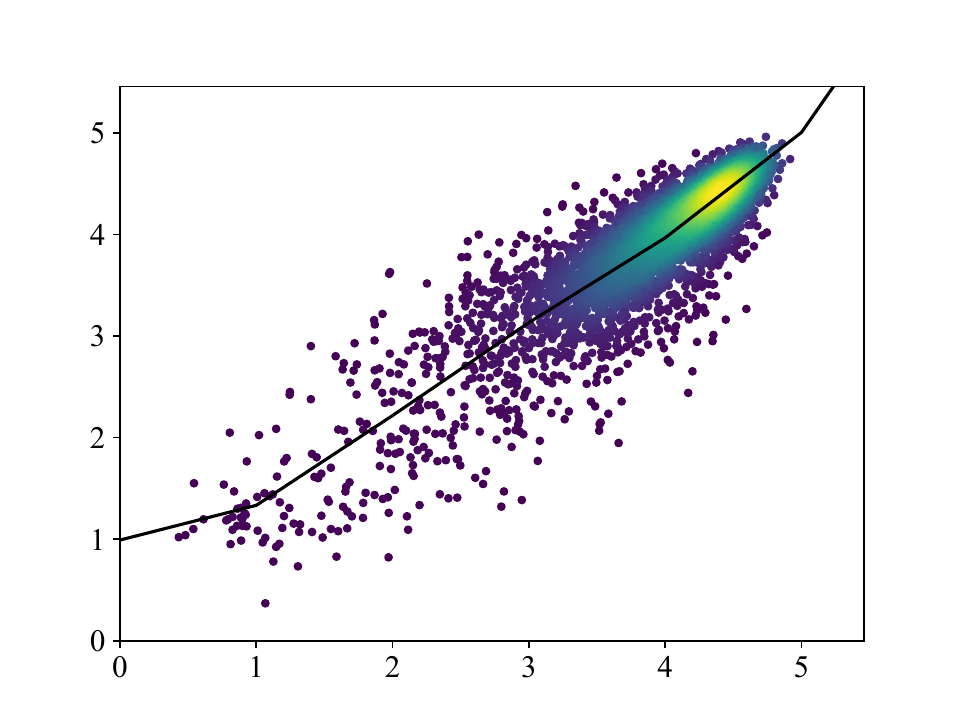}
    \caption{Oblivion}
    \label{fig:first}
\end{subfigure}
\hfill
\begin{subfigure}{0.23\textwidth}
    \includegraphics[width=\textwidth]{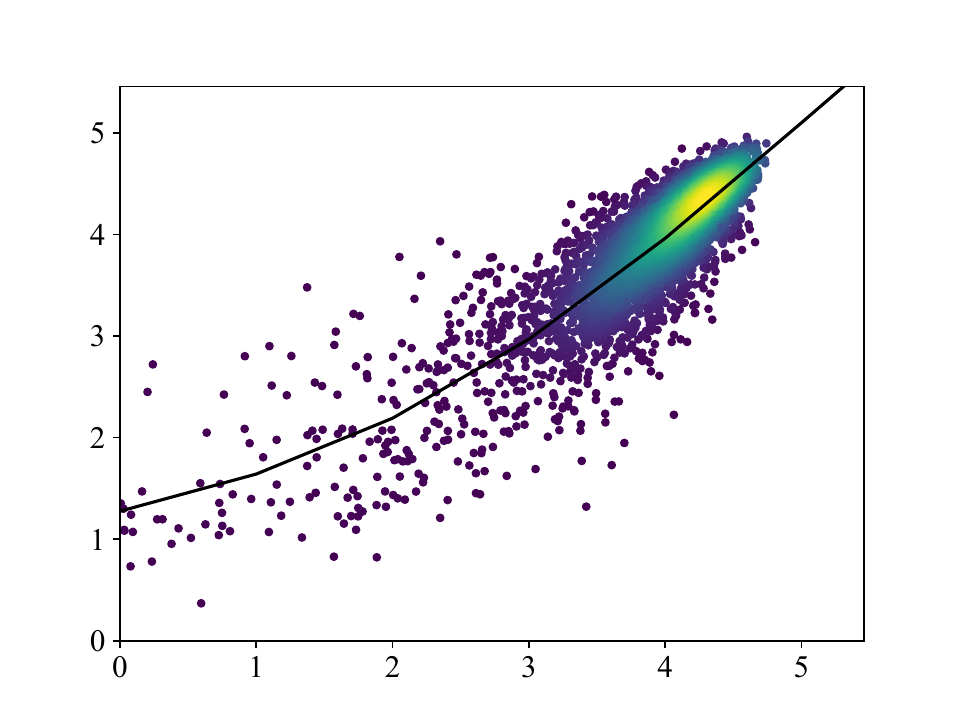}
    \caption{IVP-Lab}
    \label{fig:first}
\end{subfigure}
\caption{Scatter plots of the predicted scores vs. MOSs in the image track. The curves are obtained by a four-order polynomial nonlinear fitting.}
\label{fig:image plots}
\end{figure*}

\subsection{Challenge phases}
Both tracks consist of two phases: the developing phase and the testing phase. In the developing phase, the participants can access the generated images/videos of the training set and the corresponding prompts and MOSs. Participants can be familiar with dataset structure and develop their methods. We also release the generated images and videos of the validation set with the corresponding prompts but without corresponding MOSs. Participants can utilize their methods to predict the quality scores of the validation set and upload the results to the server. The participants can receive immediate feedback and analyze the effectiveness of their methods on the validation set. The validation leaderboard is available. In the testing phase, the participants can access the images and videos of the testing set with the corresponding prompts but without corresponding MOSs. Participants need to upload the final predicted scores of the testing set before the challenge deadline. Each participating team needs to submit a source code/executable and a fact sheet, which is a detailed description file of the proposed method and the corresponding team information. The final results are then sent to the participants.

\section{Challenge Results}
\label{Challenge Results}

\begin{figure*}[ht]
\centering
\begin{subfigure}{0.23\textwidth}
    \includegraphics[width=\textwidth]{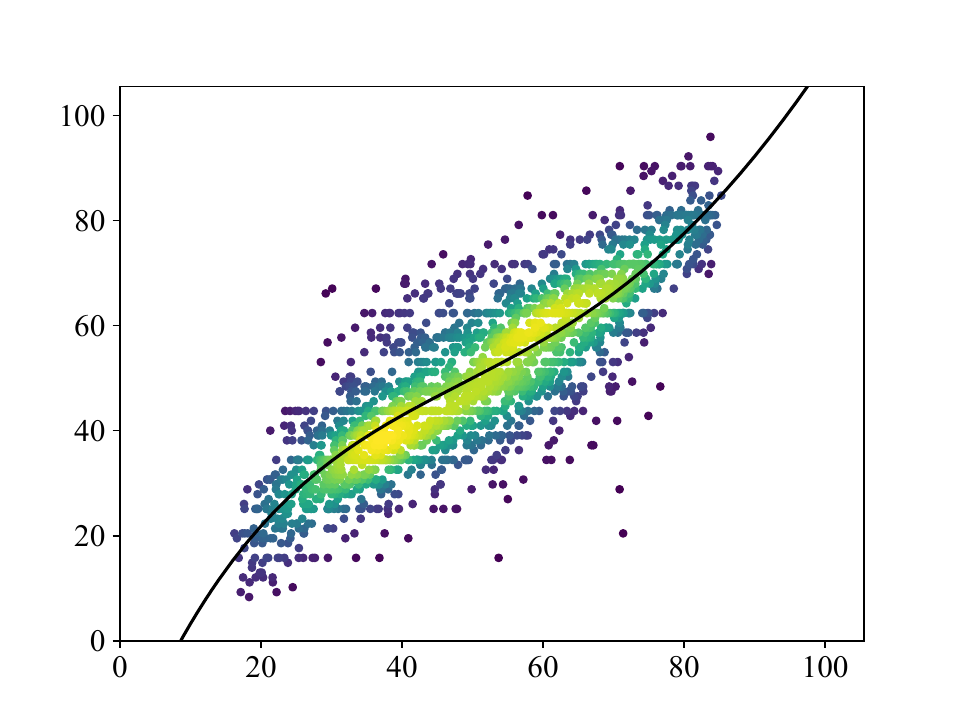}
    \caption{IMCL-DAMO}
    \label{fig:first}
\end{subfigure}
\hfill
\begin{subfigure}{0.23\textwidth}
    \includegraphics[width=\textwidth]{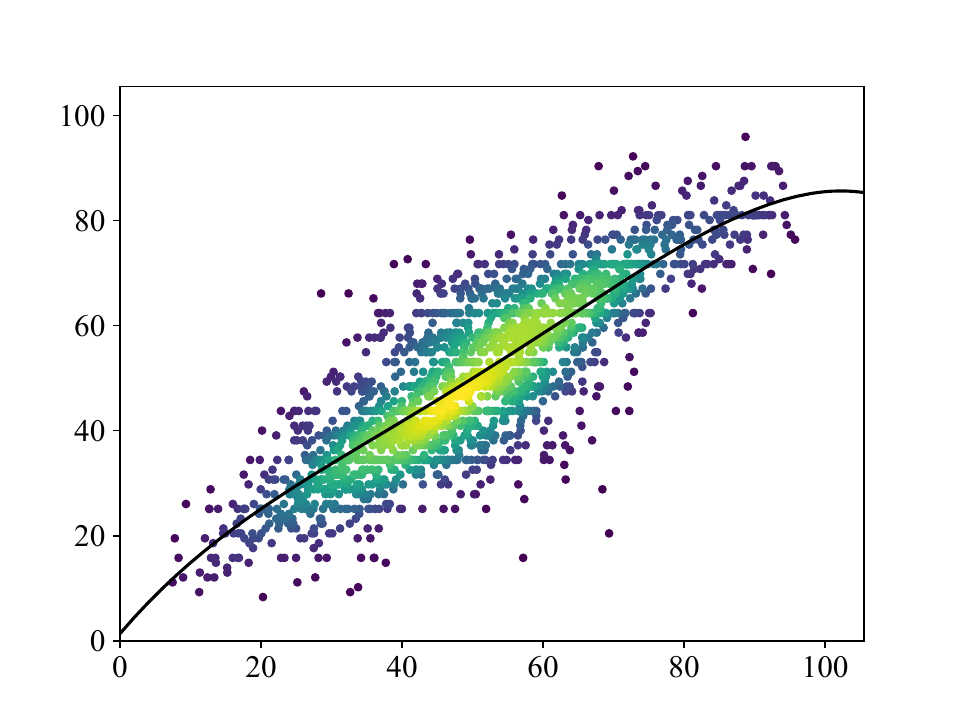}
    \caption{Kwai-kaa}
    \label{fig:first}
\end{subfigure}
\hfill
\begin{subfigure}{0.23\textwidth}
    \includegraphics[width=\textwidth]{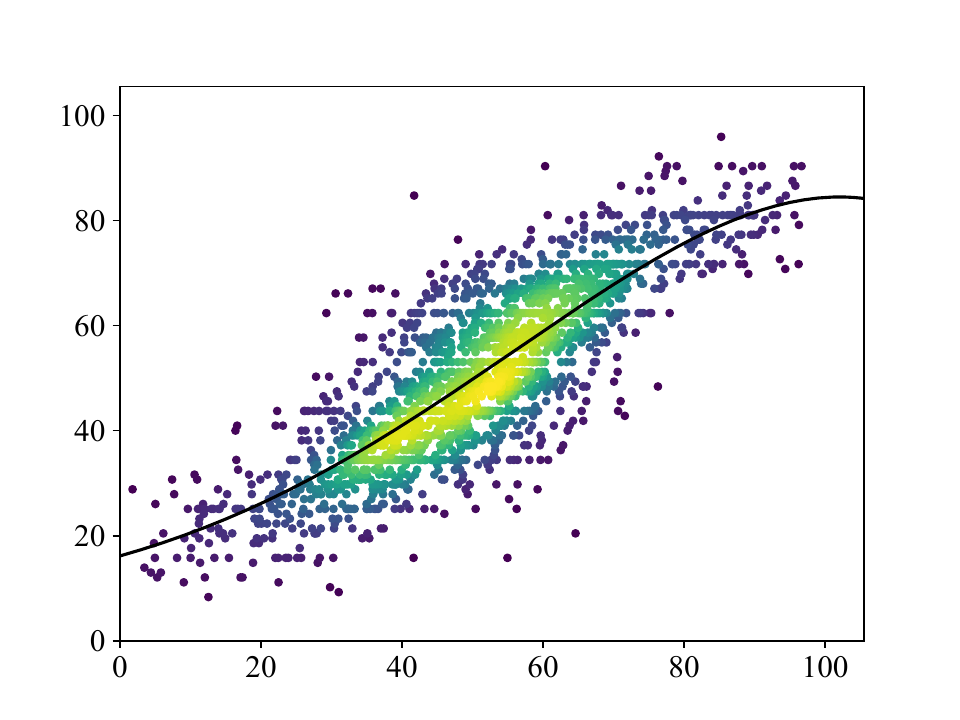}
    \caption{SQL}
    \label{fig:first}
\end{subfigure}
\hfill
\begin{subfigure}{0.23\textwidth}
    \includegraphics[width=\textwidth]{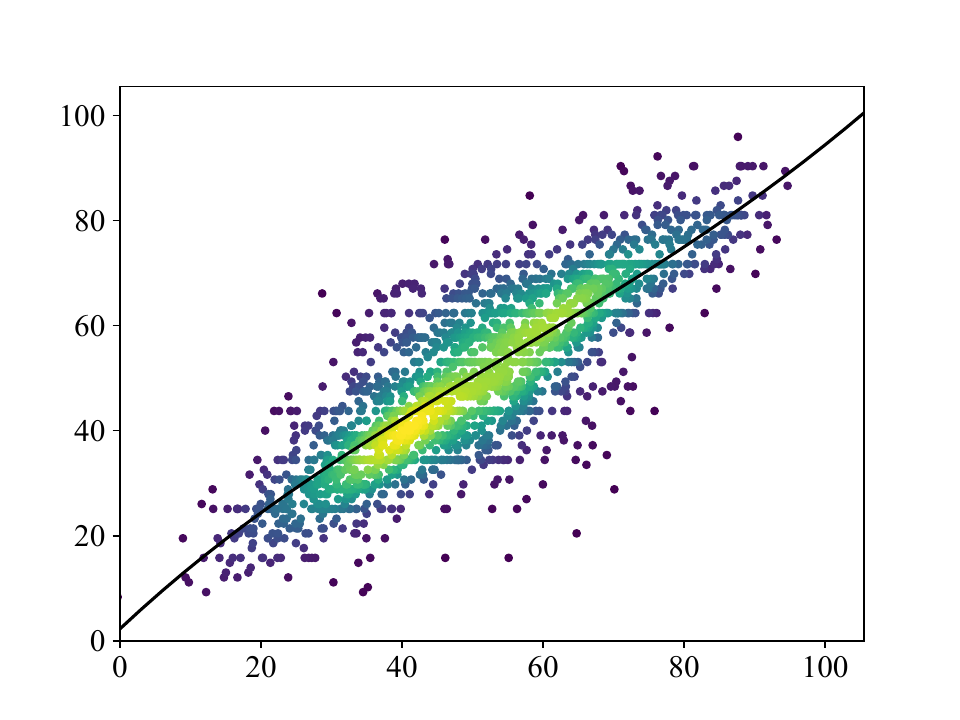}
    \caption{musicbeer}
    \label{fig:first}
\end{subfigure}
\\
\begin{subfigure}{0.23\textwidth}
    \includegraphics[width=\textwidth]{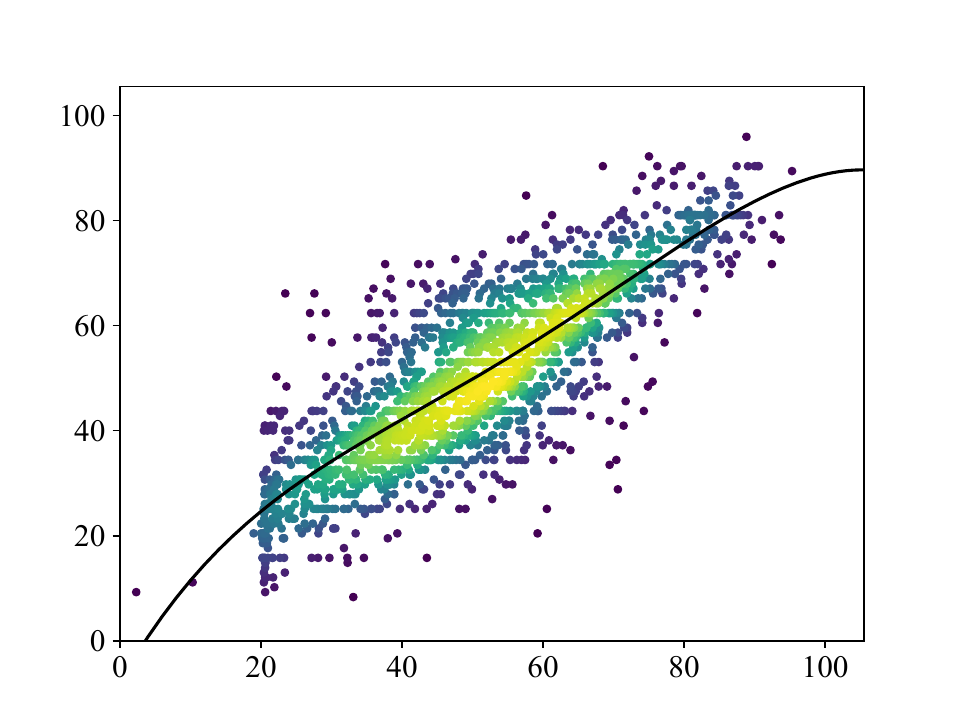}
    \caption{finnbingo}
    \label{fig:first}
\end{subfigure}
\hfill
\begin{subfigure}{0.23\textwidth}
    \includegraphics[width=\textwidth]{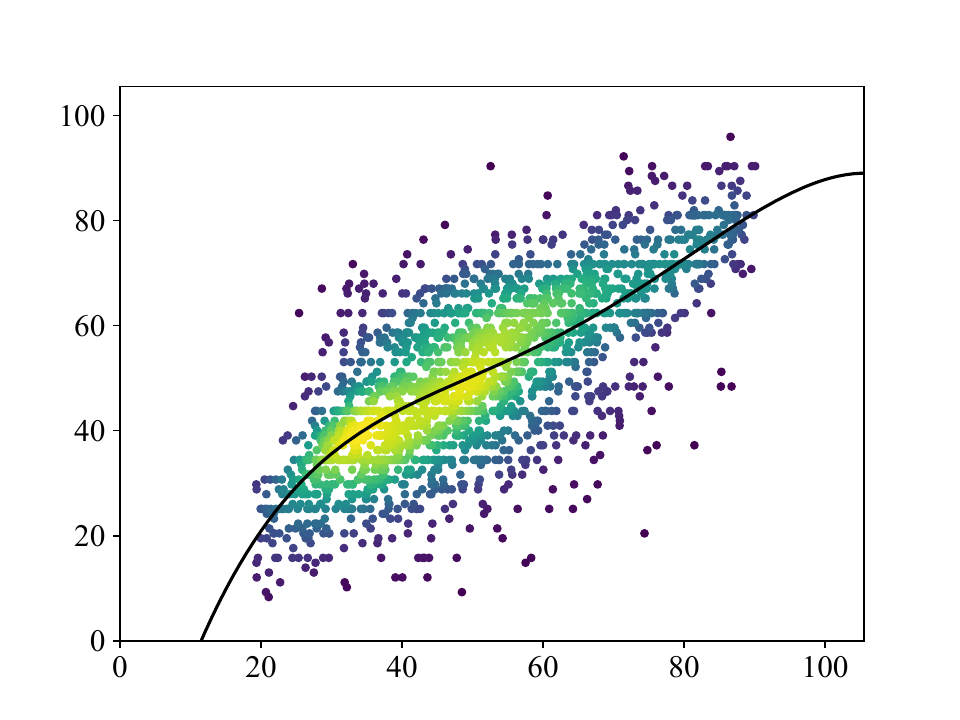}
    \caption{PromptSync}
    \label{fig:first}
\end{subfigure}
\hfill
\begin{subfigure}{0.23\textwidth}
    \includegraphics[width=\textwidth]{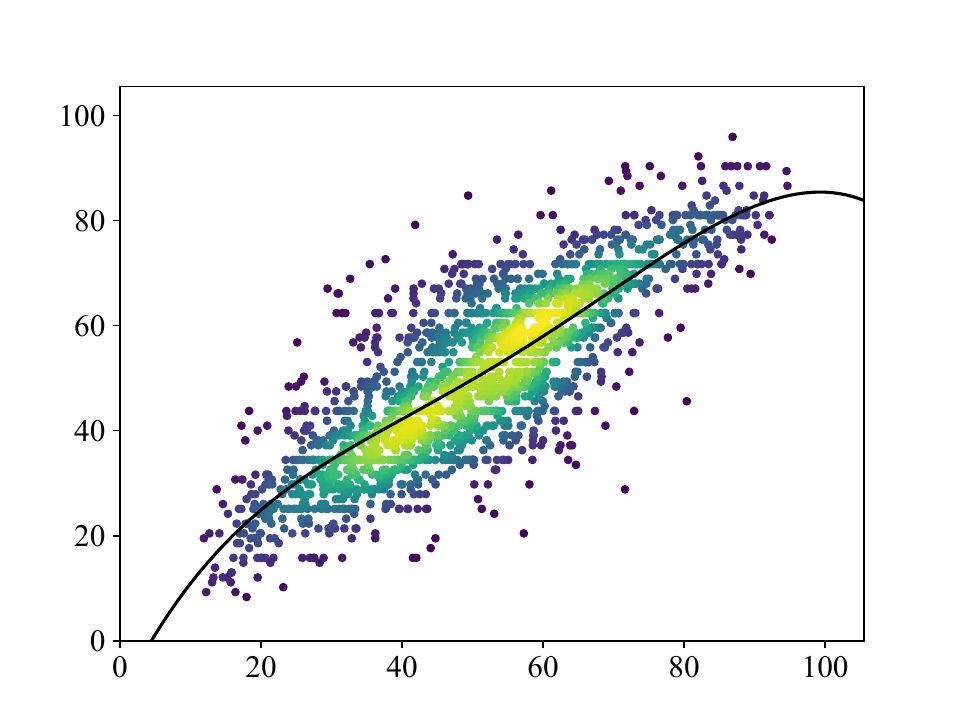}
    \caption{QA-FTE}
    \label{fig:first}
\end{subfigure}
\hfill
\begin{subfigure}{0.23\textwidth}
    \includegraphics[width=\textwidth]{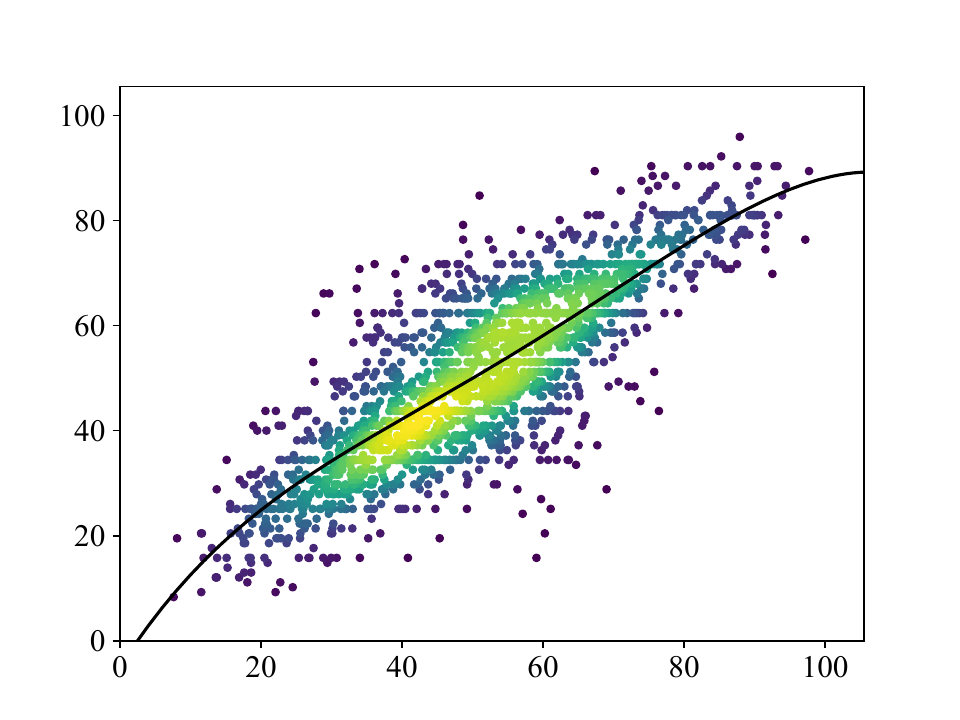}
    \caption{MediaSecurity\_SYSU\&Alibaba}
    \label{fig:first}
\end{subfigure}
\\
\begin{subfigure}{0.23\textwidth}
    \includegraphics[width=\textwidth]{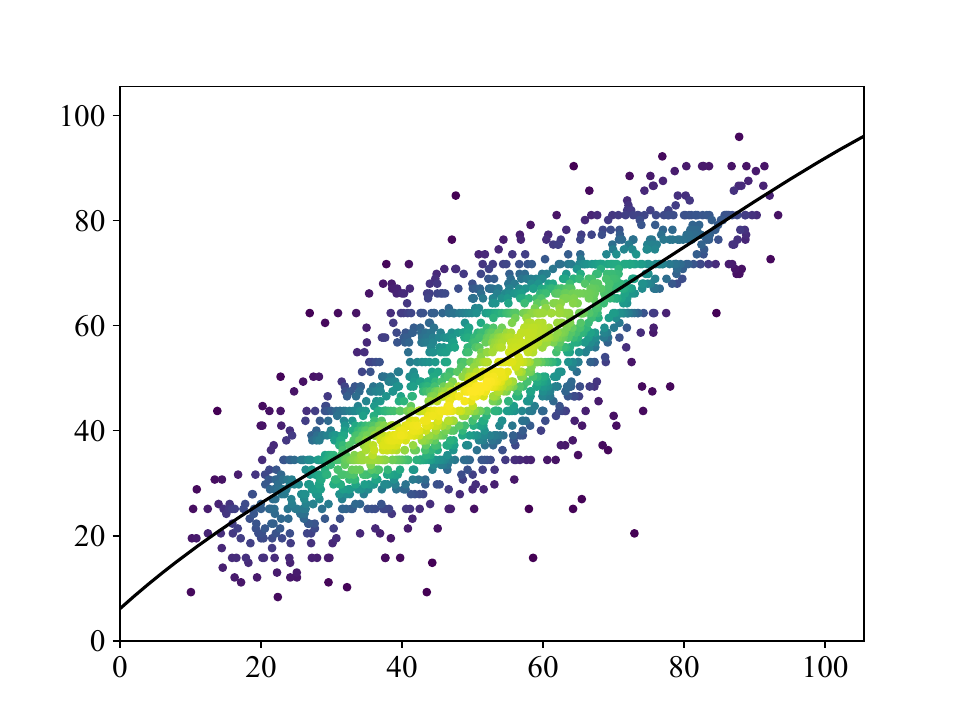}
    \caption{IPPL-VQA}
    \label{fig:first}
\end{subfigure}
\hfill
\begin{subfigure}{0.23\textwidth}
    \includegraphics[width=\textwidth]{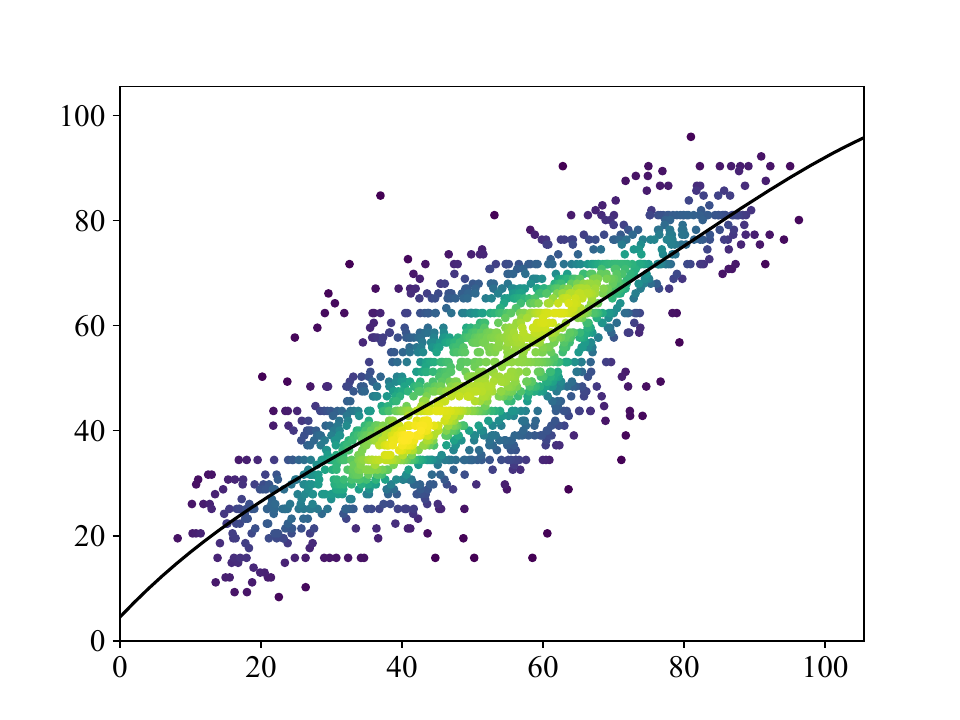}
    \caption{IVP-Lab}
    \label{fig:first}
\end{subfigure}
\hfill
\begin{subfigure}{0.23\textwidth}
    \includegraphics[width=\textwidth]{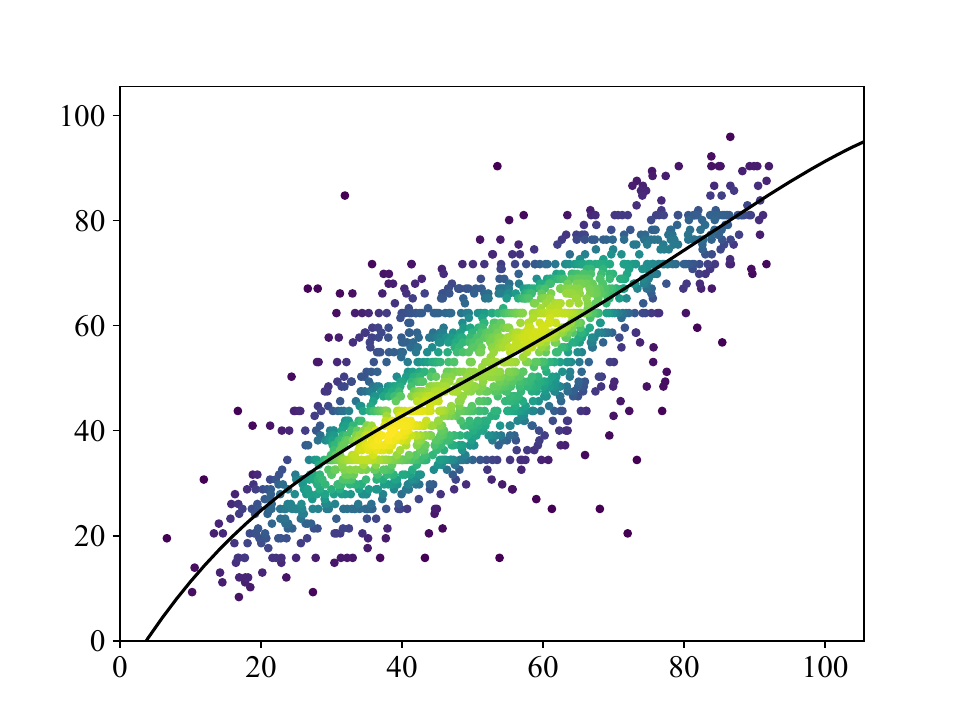}
    \caption{Oblivion}
    \label{fig:first}
\end{subfigure}
\hfill
\begin{subfigure}{0.23\textwidth}
    \includegraphics[width=\textwidth]{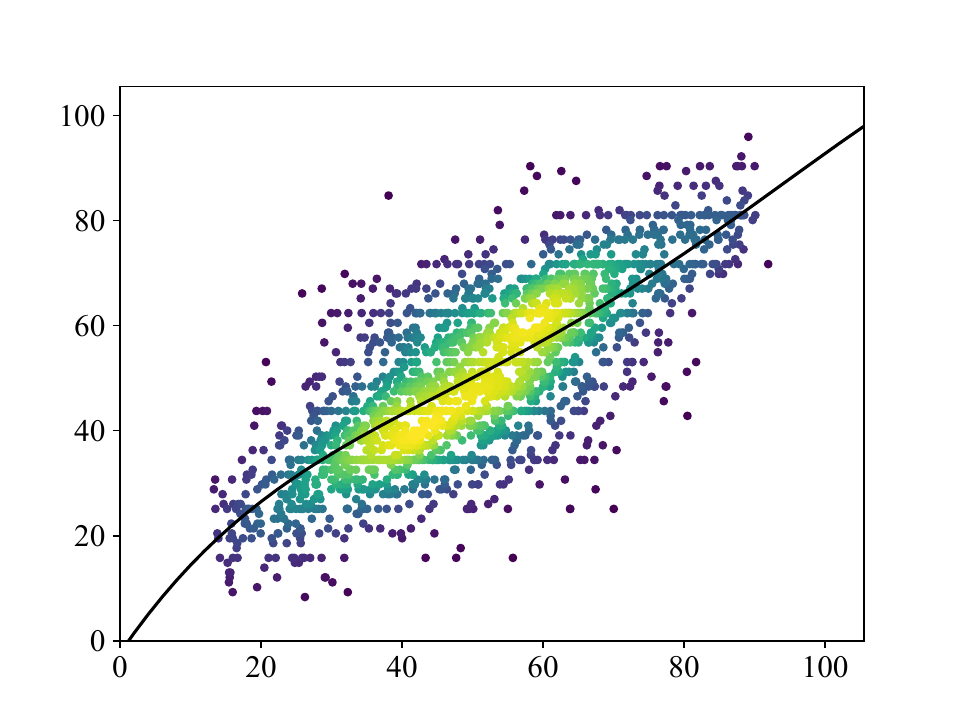}
    \caption{UBC DSL Team}
    \label{fig:first}
\end{subfigure}

\caption{Scatter plots of the predicted scores vs. MOSs in the video track. The curves are obtained by a four-order polynomial nonlinear fitting.}
\label{fig:video plots}
\end{figure*}

16 teams in the image track and 12 teams in the video track have submitted their final codes/executables and fact sheets. 
Table~\ref{tab:image results} and Table~\ref{tab:video results} summarize the main results and important information of the 28 valid teams. The methods of these teams are briefly introduced in Section \ref{Challenge Methods} and the team members are listed in Appendix \ref{sec:apd:track1team}.

\subsection{Baselines}
We compare the performance of submitted methods with several state-of-the-art I/VQA methods on the testing set, including StairIQA~\cite{li2023agiqa}, DBCNN~\cite{zhang2020blind}, and LIQE~\cite{zhang2023liqe} for the image track and SimpleVQA~\cite{sun2022deep}, FAST-VQA~\cite{wu2022fast}, and DOVER~\cite{wu2023dover} for the video track.

\subsection{Result analysis}
The main results of 28 teams' methods and the baseline methods are shown in Table~\ref{tab:image results} and Table~\ref{tab:video results}. It can be seen that in both tracks, the results of the baseline methods are not ideal in the testing set of the two datasets, while most of the submitted methods have achieved better results. It means that these methods are closer to human visual perception when used to evaluate the generated images and videos. In the image track, 9 teams achieve a main score higher than 0.9, and 4 teams are higher than 0.91. In the video track, 9 teams achieve a main score higher than 0.8, 5 teams higher than 0.82, and the championship team is higher than 0.83. In the meantime, the top-ranked teams only have a small difference in the main score. Figure~\ref{fig:image plots} and Figure~\ref{fig:video plots} show scatter plots of predicted scores versus MOSs for the 28 teams' methods on the testing set. The curves are obtained by a four-order polynomial nonlinear fitting. We can observe that the predicted scores obtained by the top team methods have higher correlations with the MOSs. They can not only meet the need to predict quality scores for generated images/videos but also contribute to improving the performance of image/video generation methods.

% most of the existing NR VQA methods are not ideal on VDPVE testing set, while the submitted methods have basically achieved good results. It means that these methods are closer to human visual perception when used to evaluate enhanced videos. $9$ teams achieve relatively better performance than FastVQA, which has good performance on the in-the-wild VQA task. Furthermore, the main scores of $4$ teams exceed $0.8$. The championship team achieves the SRCC score of $0.8576$ and the PLCC score of $0.8396$. Figure \ref{fig:Scatter plots} shows scatter plots of predicted scores versus MOSs for the 19 teams' methods on VDPVE testing set. The curves shown in Figure \ref{fig:Scatter plots} are obtained by a four-order polynomial nonlinear fitting. We can observe that the predicted scores obtained by the top team methods have higher correlations with the MOSs. They can not only meet the need to predict quality scores for enhanced videos but also contribute to improving the performance of video enhancement methods.

\section{Challenge Methods}
\label{Challenge Methods}

\subsection{Image Track}

\subsubsection{pengfei}
\label{pengfei}

\begin{figure*}
    \centering
    \includegraphics[width=\linewidth]{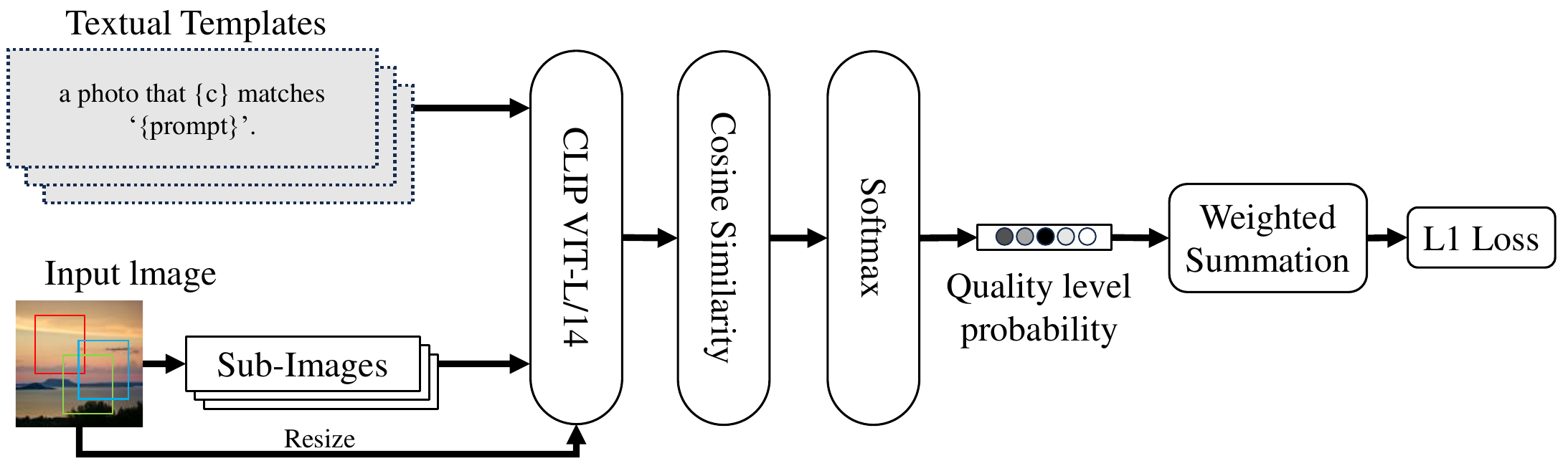}
    \caption{Overview diagram of the proposed method of team pengfei.}
    \label{fig:pengfei}
\end{figure*}

Team pengfei~\cite{pengfei} wins the championship in the image track. Their method enhances LIQE~\cite{zhang2023liqe} by considering the correlation between prompts and generated images in the AIGC task, as shown in the Figure~\ref{fig:pengfei}. To represent this correlation, they design corresponding textual templates such as ``a how image matching the prompt'', where ``how'' corresponds to five different adverbs: ``badly'', ``poorly'', ``fairly'', ``well'', and ``perfectly''. The textual templates are fed into the text encoder of the CLIP model~\cite{radford2021clip} to obtain text features. Subsequently, the images are input into the image encoder of CLIP to obtain image features. CLIP’s capability to compute the correlation between two features allows them to derive five levels of correlation between images and prompts. These levels of correlation are then weighted and summed using a softmax function to derive the final score. Additionally, considering the variation in image sizes, they not only feed image chunks into the image encoder but also include a resized version of the original image to learn global semantics.

Furthermore, in AIGI quality assessment, differences in image ratings compared to traditional image assessment ratings are observed. Traditional image assessment involves intuitive sorting based on factors like blurriness, distortion, and contamination. However, AIGI quality assessment is heavily influenced by prompts, making it difficult to effectively rank images between different prompts. Therefore, the ordinal relationship between two different image-text pairs is complex and challenging to directly learn. Traditional ranking loss is ineffective in this scenario. Instead, using L1 loss to directly fit score values and indirectly learn the order yields better SRCC scores.

Additionally, it is observed that the distribution of image ratings varies among images generated by different models. Failure to consider the model generating the images may lead to insufficient learning of the data distribution. However, considering the model generating the images may cause overfitting due to the uneven distribution of model categories. Therefore, two models are trained to integrate the results. One model undergoes normal training, while the other model has the name of the generating model added to the prompt to learn different distributions. The final inference results of the two models are multiplied and assembled.

\subsubsection{MediaSecurity\_SYSU\&Alibaba}

Team MediaSecurity\_SYSU\&Alibaba wins second place in the image track. They use both single-modal and multiple multi-modal networks for learning. The single-modal model used the backbone network of EVA02 large~\cite{EVA02}, while
the multi-modal model used EVA02 large, ConvNeXt~\cite{liu2022convnet}, and ConvNeXt v2~\cite{woo2023convnext} as the
image branches of the backbone network. They were combined with the text branches of the backbone network, such as $\text{Bert}_{base}$~\cite{devlin2018bert}, $\text{RoBERTa}_{base}$~\cite{liu2019roberta}, and $\text{DeBERTaV3}_{base}$~\cite{he2021debertav3}, to learn and evaluate scores. The training is done either in the \engordnumber{1} or \engordnumber{2} fold of the 10-fold training, and finetuned across the entire training set. They use a total of 20 models, and the detailed combinations
are shown below.

- weight=32/3200, convnext\_large\_in22k~\cite{liu2022convnet} + bert-base-uncased~\cite{devlin2018bert} (max\_length=64), VisualBert~\cite{li2019visualbert} (trained in fold 0, num\_fold=10)

- weight=48/3200, convnext\_large\_in22k~\cite{liu2022convnet} + bert-base-uncased~\cite{devlin2018bert} (max\_length=64), VisualBert~\cite{li2019visualbert} (finetuned in all training set)

- weight=32/3200, convnext\_xlarge\_in22kk~\cite{liu2022convnet} + bert-base-uncased (max\_length=32)~\cite{devlin2018bert}, VisualBert~\cite{li2019visualbert} (trained in fold 0, num\_fold=10)

- weight=48/3200, convnext\_xlarge\_in22k~\cite{liu2022convnet} + bert-base-uncased~\cite{devlin2018bert} (max\_length=32), VisualBert~\cite{li2019visualbert} (finetuned in all training set)

- weight=32/3200, convnext\_large\_in22k~\cite{liu2022convnet} + deberta-v3-base~\cite{he2021debertav3} (max\_length=64), concatenate features (trained in fold 0, num\_fold=10)

- weight=48/3200, convnext\_large\_in22k~\cite{liu2022convnet} + deberta-v3-base~\cite{he2021debertav3} (max\_length=64), concatenate features (finetuned in all training set)

- weight=32/3200, convnext\_large\_in22k~\cite{liu2022convnet} + roberta-base~\cite{liu2019roberta} (max\_length=64), concatenate features (trained in fold 0, num\_fold=10)

- weight=48/3200, convnext\_large\_in22k~\cite{liu2022convnet} + roberta-base~\cite{liu2019roberta} (max\_length=64), concatenate features (finetuned in all training set)

- weight=32/3200, convnext\_xlarge\_in22k~\cite{liu2022convnet} + \seqsplit{deberta-v3-base}~\cite{he2021debertav3} (max\_length=64), concatenate features (trained in fold 0, num\_fold=10)

- weight=48/3200, convnext\_xlarge\_in22k~\cite{liu2022convnet} + \seqsplit{deberta-v3-base}~\cite{he2021debertav3} (max\_length=64), concatenate features (finetuned in all training set)

- weight=50/3200, \seqsplit{convnextv2\_huge.fcmae\_ft\_in22k\_in1k\_512}~\cite{woo2023convnext} + bert-base-uncased~\cite{devlin2018bert} (max\_length=75), concatenate features (trained in fold 0, num\_fold=10)

- weight=75/3200, \seqsplit{convnextv2\_huge.fcmae\_ft\_in22k\_in1k\_512}~\cite{woo2023convnext} + bert-base-uncased~\cite{devlin2018bert} (max\_length=75), concatenate features (finetuned in all training set)

- weight=560/3200, \seqsplit{eva02\_large\_patch14\_448.mim\_m38m\_ft\_in22k\_in1k}~\cite{EVA02} + deberta-v3-large~\cite{he2021debertav3} (max\_length=75) + FC + LeakyReLU, concatenate features (trained in fold 0, num\_fold=10)

- weight=1040/3200, \seqsplit{eva02\_large\_patch14\_448.mim\_m38m\_ft\_in22k\_in1k}~\cite{EVA02} + deberta-v3-large~\cite{he2021debertav3} (max\_length=75) + FC + LeakyReLU, concatenate features (finetuned in all training set)

- weight=275/3200, \seqsplit{eva02\_large\_patch14\_448.mim\_m38m\_ft\_in22k}~\cite{EVA02} + bert-base-uncased~\cite{devlin2018bert} (max\_length=75), concatenate features (trained in fold 2, num\_fold=10)

- weight=325/3200, \seqsplit{eva02\_large\_patch14\_448.mim\_m38m\_ft\_in22k}~\cite{EVA02} + bert-base-uncased~\cite{devlin2018bert} (max\_length=75), concatenate features (finetuned in all training set)

- weight=50/3200, \seqsplit{eva02\_large\_patch14\_448.mim\_m38m\_ft\_in22k}~\cite{EVA02}  + bert-base-uncased~\cite{devlin2018bert} (max\_length=75) + decoder$*6$, concatenate features (trained in fold 2, num\_fold=10)

- weight=87.5/3200, \seqsplit{eva02\_large\_patch14\_448.mim\_m38m\_ft\_in22k}~\cite{EVA02} + bert-base-uncased (max\_length=75)~\cite{devlin2018bert} + decoder$*6$, concatenate features (finetuned in all training set)

- weight=50/3200, \seqsplit{eva02\_large\_patch14\_448.mim\_m38m\_ft\_in22k}~\cite{EVA02} + roberta-base (max\_length=75)~\cite{liu2019roberta}, concatenate features (trained in fold 2, num\_fold=10)

- weight=87.5/3200, \seqsplit{eva02\_large\_patch14\_448.mim\_m38m\_ft\_in22k}~\cite{EVA02} + roberta-base~\cite{liu2019roberta} (max\_length=75), concatenate features (finetuned in all training set)

\subsubsection{geniuswwg}
\label{musicbeer}

\begin{figure}
    \centering
    \includegraphics[width=\linewidth]{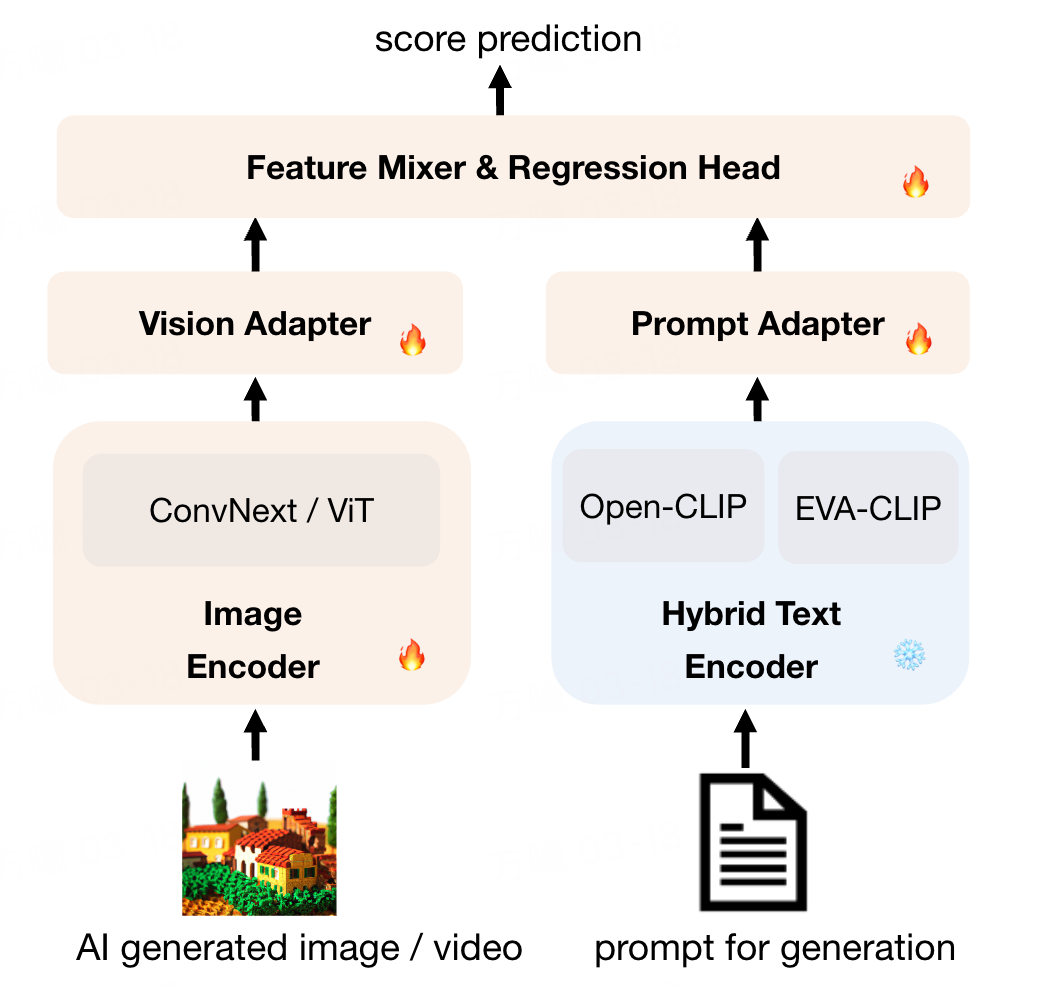}
    \caption{Overview of team geniuswwg proposed method.}
    \label{fig:musicbeer}
\end{figure}

\begin{figure*}[t]
    \centering
    \includegraphics[width=\linewidth]{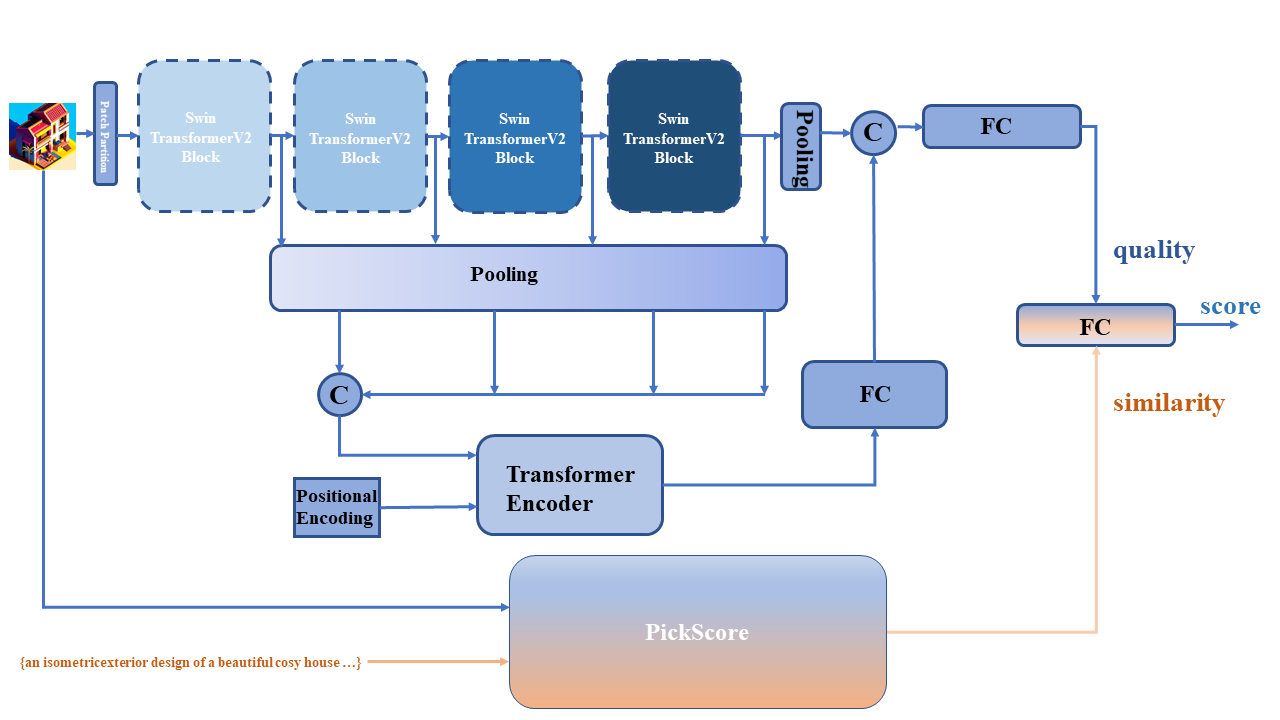}
    \caption{Overview of team Yag proposed method.}
    \label{fig:yag}
\end{figure*}

Team geniuswwg~\cite{fang2024pcqa} wins third place in the image track. They propose an innovative approach to assess the quality of AIGC by treating it as a regression task under specified prompt conditions. It employs a dual-source CLIP~\cite{radford2021clip} text encoder to interpret prompts, combining their features for a nuanced understanding. Image features are extracted using models like ConvNeXt~\cite{liu2022convnet}, pre-trained on ImageNet~\cite{deng2009imagenet}, and adapted for interaction with the text features and vision features. A feature mixer module then blends the text and video features, using dot product and concatenation to model their correlation and conditional relationship. The final quality score is predicted by a two-layer Multi-Layer Perception (MLP). To enhance generalization, the system applies light data augmentations like flips and brightness adjustments. The ensemble method further refines the assessment by blending predictions from three diverse models, normalized for consistency, and averaged to produce a robust quality evaluation. Figure~\ref{fig:musicbeer} shows the overview of their method.

Concretely, they use the frozen CLIP text encoder to encode the prompt, whose pre-trained weights are brought from two different open-source CLIP implements (\ie Open-CLIP~\cite{radford2021clip} and EVA-CLIP~\cite{sun2023evaclip}) and pre-trained on different datasets (\eg DFN-5B~\cite{fang2023dfn5b}, LAION-2B~\cite{schuhmann2022laion}, DataComp-1B~\cite{gadre2024datacomp} and WebLI). They build a hybrid prompt encoder by simply concatenating the output features from these two different CLIP text encoders. After obtaining the text features for prompts, they use a trainable dense layer as a prompt adapter, to align features to make better interaction with image features.

They simply use ConvNeXt-Small~\cite{liu2022convnet} (or ViT~\cite{dosovitskiy2020vit} and any other backbones with ImageNet~\cite{deng2009imagenet} pre-trained weights) as the vision backbones. Similarly, they use a trainable dense layer as a vision adapter.

After the adapted prompt feature and vision features have been obtained, they use a module named feature mixer to make these two features interact with each other. They propose to use the features of prompts as a condition. They introduce two types of lightweight feature mixers: dot product and concatenation. The dot product can more effectively model the correlation between the generated images and the prompt. The use of concatenation is more akin to treating the prompt as a conditional factor. They employ these two types of feature mixers with different experts and ultimately utilize them for model blending. After obtaining the fused features, they employ a two-layer MLP as the prediction head to regress the final quality score.

Besides, they use random horizontal flip, slight random resized crop, and slight brightness contrast transformation for augmentation. These augmentations are relatively minor and generally do not affect the subjective quality assessment of the image, but can improve the model's generalization ability.

They blend 3 different models with different vision backbones: ConvNeXt~\cite{liu2022convnet}, ViT-Transformer~\cite{dosovitskiy2020vit}, and EVA02-Transformer~\cite{EVA02}. Firstly, we normalize the predicted scores of each model on the testing set, so that different models have the same mean and variance of prediction. Finally, they blend all the models by averaging. 

\subsubsection{Yag}

Team Yag~\cite{Yu2024ntire} feeds images sampled through SAMA~\cite{liu2024scaling} into Swin Transformer V2~\cite{liu2022swinv2} for the image quality prediction component. Swin Transformer V2 is adept at extracting local features of video frames across various levels. By normalizing and rescaling the local features via a feature pooling operation, they obtain global frame-level features through the amalgamation of local representations across a multi-scale fusion module, which comprises a series of transformer layers. Subsequently, they concatenate the local and global features along the channel dimension and input them into a linear layer to ascertain the quality score, which is inspired by TReS~\cite{golestaneh2022no}. This methodology harnesses spatial and temporal information from both global and local perspectives, thereby enhancing the perceptual capability of video quality assessment.

For the image-text similarity prediction component, they employ PickScore~\cite{kirstain2024pick} to predict the similarity between images and text. They input the results from both the quality prediction and similarity prediction into a fully connected layer to derive the final quality score. The overview of the proposed method is shown in Figure~\ref{fig:yag}.

Besides the provided training data, they also use CLIVE~\cite{ghadiyaram2015clive}, LIVE~\cite{sheikh2005live}, KonIQ-10K~\cite{hosu2020koniq}, KADID-10K~\cite{lin2019kadid}, AGIQA-1k~\cite{zhang2023perceptual}, AGIQA-3K~\cite{li2023agiqa}, AIGCIQA2023~\cite{wang2023aigciqa2023} and PKU-I2IQA~\cite{yuan2023pku} as additional data. The training images are paired-cropped into $256 \times 256$ patches for the image quality prediction component and $224 \times 224$ patches for the image-text similarity prediction component. They train the model using the Adam optimizer, setting the initial learning rate to $2e^{-5}$ for the Swin Transformer V2~\cite{liu2022swinv2}, to $2e^{-6}$ for AIGC images, and to $1e^{-6}$ for the entire model.

\subsubsection{QA-FTE}
\label{qafte}

\begin{figure*}
    \centering
    \includegraphics[width=\linewidth]{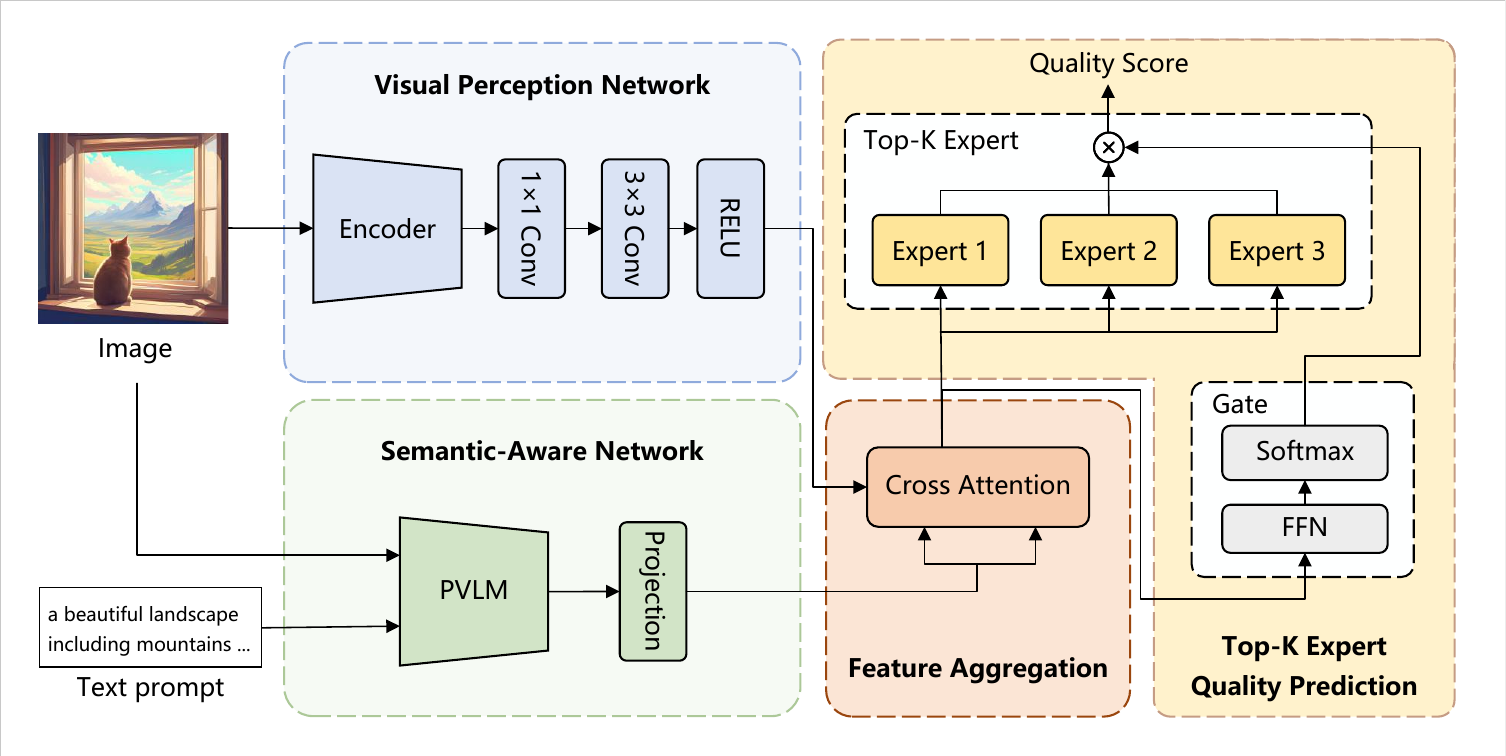}
    \caption{Overview of team HUTB-IQALab proposed method.}
    \label{fig:hutb}
\end{figure*}

Team QA-FTE proposes a vision-language fused video quality evaluator, which is designed for AIGC. Considering the quality of AI-generated images is affected by the consistency of vision and language, they use CLIP~\cite{radford2021clip} as the backbone model. Specifically, they first use the CLIP encoder to extract the vision feature of the image and the language feature of the prompt. Then, a bilinear pooling is used to obtain an interactive feature, which represents the consistency between vision information and language information. Finally, the above features are fused to predict AI-generated image quality scores.

\subsubsection{HUTB-IQALab}

Team HUTB-IQALab~\cite{yang2024moeagiqa} proposes a novel mixture-of-experts boosted visual perception and semantic-aware quality assessment for AI-generated images. Firstly, they design the visual perception network to establish perceptual rules to obtain visual perception features. Secondly, they enhance the diversity of degradation-specific knowledge through the semantic-aware network, generating semantic-aware features. Thirdly, instead of fusing on predicted image scores, they propose to conduct cross-attention on visual perception and semantic-aware features, so that they can obtain comprehensive features and the inherent correlation between these features. Finally, they propose a mixture-of-experts model, involving
multiple experts working collaboratively. Each expert is responsible for a
specific set of features and outputs a corresponding prediction score. The
mixture of multiple experts will ultimately yield a holistic, perceptually-aware score. Figure~\ref{fig:hutb} shows the overview of the proposed method.

\subsubsection{IQ Analyzers}

IQ Analyzers propose a methodology that adopts a Mixture-of-Experts approach, integrating a broad spectrum of feature types. This includes low-level quality features obtained from Re-IQA~\cite{saha2023re}, text-to-image alignment features via BLIP2~\cite{li2023blip2} and ImageBind~\cite{girdhar2023imagebind}, image aesthetics representations from VILA~\cite{ke2023vila}, the naturalness attributes of images as judged by DINOv2~\cite{oquab2023dinov2}, and traits from the text-to-image human preference model, ImageReward~\cite{xu2024imagereward}. The features are combined and utilized to train an ElasticNet model, which maps the aggregated representation to the MOSs.

\subsubsection{PKUMMCAL}

\begin{figure*}
    \centering
    \includegraphics[width = \linewidth]{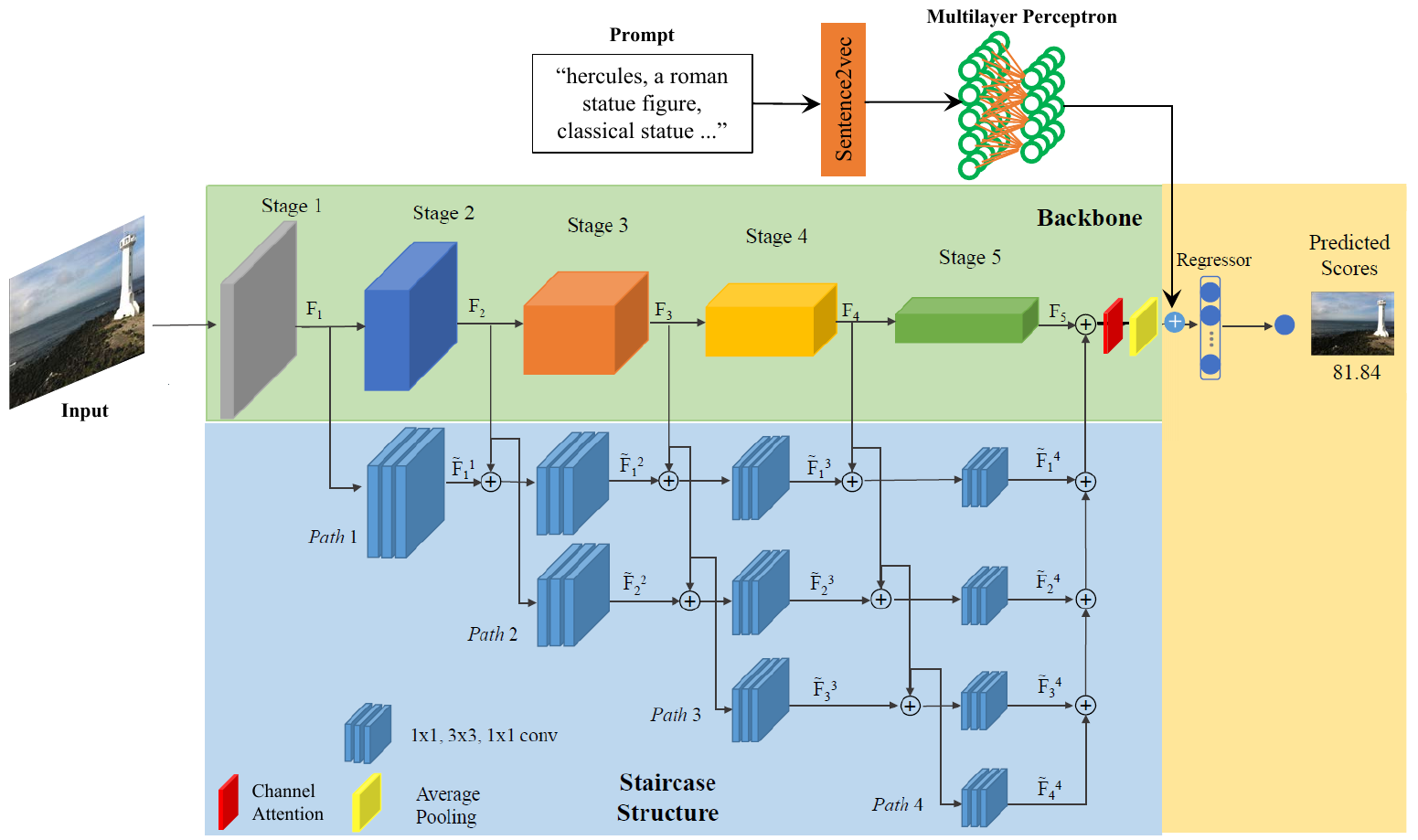}
    \caption{Overview of team JNU\_620 proposed method.}
    \label{fig:jnu620}
\end{figure*}

Team PKUMMCAL believes that compared to the task of Natural Scene Image Quality Assessment (NSIQA), which focuses only on the perceptual quality of images, the quality evaluation of AI-generated images needs to consider the text-image consistency additionally. Therefore, they bring textual information into the model using a text encoder pre-trained in CLIP~\cite{radford2021clip}. In terms of training methods, inspired by the works on natural image quality evaluation based on multi-task learning, they introduce additional tasks, hoping that the model can learn auxiliary knowledge from them. Unlike natural images, the distortion types of AI-generated images are difficult to distinguish directly, so they choose to predict the generative model used for creating AI-generated images as their auxiliary task. At the same time, to utilize the consistency of text and image information, they integrate the fine-tuning method from CLIPIQA~\cite{wang2023exploring} into their model. They specifically designed three models, which are ResNet50-based~\cite{he2016deep}, DINOv2-based~\cite{oquab2023dinov2}, and ConvNeXt-Based~\cite{liu2022convnet}. Besides, they integrate the HyperNet part of HyperIQA~\cite{Su_2020_CVPR} into the ResNet50-based model to achieve semantic adaptive evaluation for different images.

The three models all share a similar framework, which is a dual-stream architecture to simultaneously process the image and its associated textual prompt. To fusion the visual and textual information, they propose an attention-based module. The fused feature and the visual feature are both fed to the quality prediction head, while the visual feature is also fed to the generation model classification head. Meanwhile, there is something different in ResNet50-based model architecture. They integrate several effective modules proposed for the NSIQA task, as illustrated.

\subsubsection{BDVQAGroup}

Team BDVQAGroup chooses two methods to assess the quality of AIGIs. One is Q-Align~\cite{wu2023qalign}, which is based on large multi-modality models
(LMMs). Q-Align converts MOSs into rating levels and uses classification to teach LMMs with text-defined rating levels instead of scores. During inference, it extracts the close-set probabilities of rating levels and performs a weighted average to obtain the LMM-predicted score. Another is based on MSTRIQ~\cite{wang2022mstriq}, a Swin-Transformer based
method. They use several data augmentation methods to increase the training dataset and enhance the robustness of MSTRIQ, which are: 1)  Expand the image along its longer side to form a square. 2) Randomly rotate the image at a degree of 90. 3) Resize the image to $448 \times 448$. 4) Randomly resize and crop the image at a ratio of 0.7. 

They use a Q-Align model pre-trained on KonIQ~\cite{hosu2020koniq}, SPAQ~\cite{fang2020spaq}, KADID~\cite{lin2019kadid}, AVA~\cite{gu2018ava}, and LSVQ~\cite{ying2021patch}, and finetune this model through three strategies. The first model is based on the Q-Align Image Quality Scorer, which is finetuned for 4 epochs on the AGIQA-1K~\cite{zhang2023perceptual} images and then finetuned for another 2 epochs on the provided training images. The second model is based on the Q-Align Image Aesthetic Scorer, which is also finetuned for 2 epochs on the provided training images. The third model is based on the Q-Align Image Quality Scorer, which is finetuned for 2 epochs on the provided training images. The fourth model is an MSTRIQ model pre-trained on TID2013~\cite{ponomarenko2015image}, KonIQ-10k~\cite{hosu2020koniq} and PIPAL~\cite{jinjin2020pipal}, and finetune this model for 150 epochs on the provided training images.

The following methods are implemented to increase model performance: 
1) Expand the image along its longer side to form a square. 2) Randomly crop the image 18 times. The crop size is $384 \times 384$ for all the 18 patches. 3) They use four models to predict and weight the four prediction results according to the following weights to obtain the final output. That is:

\begin{equation}
\begin{split}
    output =  &(0.3 \times model_1 + (0.4 \times model_2 + 0.6 \times  \\
    &model_3) \times 0.7) \times 0.8 + 0.2 \times model_4.
\end{split}
\end{equation}

\subsubsection{JNU\_620}

Team JNU\_620 designs a method based on StairIQA~\cite{sun2023blind}, which includes two parts, a staircase network, and an image quality regressor. To make the model pay more attention to the important features, they added channel attention at the end of the staircase network. Moreover, to make full use of the prompts, they leverage the Sentence2Vec technology to convert prompts into sentence vectors. Then, the sentence vectors are transferred into the multi-layer perceptron, which strengthens the representation of the features. After extracting prompt-aware features by the multi-layer perceptron and quality-aware features by the staircase network, they add these features and map them to the quality scores with a regression model. Figure~\ref{fig:jnu620} shows the overview of the proposed method. 

They use seven models as the backbone of the proposed model, including ShuffleNet~\cite{zhang2018shufflenet}, MobileNetV2~\cite{sandler2018mobilenetv2}, MobileNetV3~\cite{howard2019mobilenetv3}, ResNet50~\cite{he2016deep}, Res2Net50~\cite{gao2019res2net}, ResNeXt50~\cite{xie2017aggregated}, and ResNeSt~\cite{zhang2022resnest}. The weights of the backbone are initialized by training on ImageNet~\cite{deng2009imagenet}, and other weights are randomly initialized. The proposed model is only trained on the provided image training set. In the training stage, images are resized to $680 \times 680$ and randomly cropped with resolutions of $640 \times 640$. The Adam method is employed for optimization with $\beta_1 = 0.9$ and $\beta_2 = 0.999$. The initial learning rate was set to $3 \times 10^{-3}$. MSE loss is used as the loss function for training. In the testing phase, the seven models are used for the ensemble. By performing the model ensemble, the results produced by multiple models are averaged for better results.

\subsubsection{MT-AIGCQA}

Similar to VBench~\cite{huang2024vbench}, team MT-AIGCQA integrates a variety of basic models by fine-tuning or directly testing on the provided dataset to obtain image quality scores in different dimensions. The basic models include improved versions of BLIP~\cite{li2022blip}, CLIP~\cite{radford2021clip}, etc. The inspection dimensions include image-text consistency, image quality, etc. Finally, they fuse the scores of the basic models to obtain the final MOS.

Specifically, for the image quality dimension, they use three models to obtain deep differential information, based on ResNet~\cite{he2016deep}, NAS~\cite{zoph2016neural}, and Swin Transformer~\cite{liu2021swin} respectively. Since there is no separate image quality score in the training set, the MOS is expressed as the target image quality score. For the image-text consistency dimension, they obtaine the image-text consistency score by finetuning BLIP~\cite{li2022blip} on the training set. Since there is no separate image-text consistency score in the training set, they regard image-text pairs with MOSs exceeding 2.5 (ranging from 0 to 5) as matches, and MOSs lower than 2.5 as mismatches. For the overall image quality dimension, they added an MLP layer to BLIP for the regression of MOS.

In the testing phase, they use the five models trained in the previous stage to obtain the corresponding basic scores respectively, and then use a series of pre-trained models to obtain the corresponding scores of image-text pairs, including CLIP~\cite{radford2021clip}, Q-Instruct~\cite{wu2023qinstruct} and ImageReward~\cite{xu2024imagereward}. Following~\cite{huang2024t2i}, they obtain the corresponding BLIP score and CLIP score. Finally, for images generated by different models, they use a linear regression model to fit the final output MOS through the scores of the basic models.

\subsubsection{IVL}

Team IVL exploits BLIP-2~\cite{li2023blip2} to encode prompt and image, respectively. BLIP-2 consists of a vision encoder, a language model, and a Querying Transformer (Q-Former). The input image is first resized to the resolution of $224 \times 224$ pixels and then fed to the model that outputs a feature map of $32 \times 768$ features. Spatial features are
finally averaged to obtain the 768-dimensional feature vector. The text prompt is first tokenized and then fed to the model which outputs a
feature map with shape $12 \times 768$. Following~\cite{li2023blip2}, they select the first token as representative of the whole text input. The two feature vectors are l2-normalized and concatenated into a 1536-dimensional feature vector. A Support Vector Regression (SVR) machine with a Radial Basis Function (RBF) kernel is used to map the features into the final quality score.

\subsubsection{CVLab}

Team CVLab proposed model is based on a pre-trained text encoder (CLIP~\cite{radford2021clip}) and an image encoder (ConvNeXt~\cite{liu2022convnet}). They use the image encoder to extract features from the images, and then pass those features with the dropout function (ratio 0.3) to a full-connection layer with 1000 input and 512 output. At the same time, they use the text encoder and the tokenized prompts, to extract the text features. Next, they concatenate the image features and text features and provide a vector with 1024 dimensions. They use this concatenated image-text feature to predict the final MOS with another fully connected layer.

\begin{figure*}[t]
    \centering
    \includegraphics[width=\linewidth]{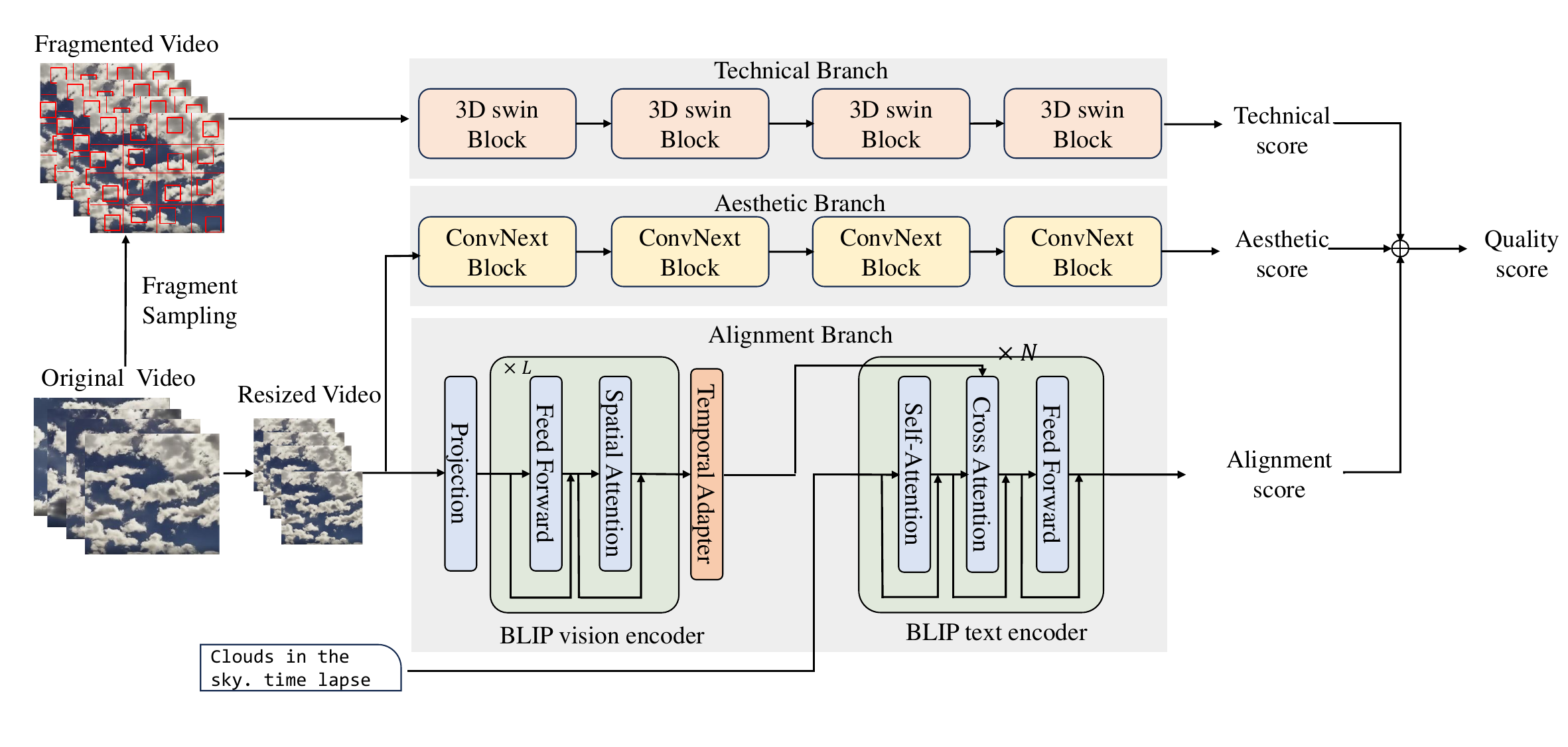}
    \caption{The overview of IMCL-DAMO team proposed versatile video quality evaluator.}
    \label{fig:imcl}
\end{figure*}

\subsubsection{z6}

Team z6 introduces a network that combines image and text features for assessing the quality of AI-generated images. The network comprises three main components. The initial component is the image feature encoder, inspired by the architecture of the staircase network~\cite{sun2023blind}. For the text feature encoder, they employ a basic transformer network. Finally, the feature fusion component utilizes a concatenation function along with a straightforward $1 \times 1$ convolution operation.

\subsubsection{Oblivion}

Team Oblivion uses a Swin Transformer~\cite{liu2021swin} as the visual backbone, and then they use CLIP~\cite{radford2021clip} to get the text and visual feature relationship. These features are used to enhance visual understanding to evaluate the quality of the image, and a Densenet~\cite{huang2017densely} is used to help it get a high visual understanding. They use the Swin-tiny network pre-trained on the Kinetics-400~\cite{kay2017kinetics} dataset to initialize the Swin Transformer backbone, and the ViT~\cite{dosovitskiy2020vit} pre-trained on the Kinetics-400 dataset to initialize the CLIP model.

\subsubsection{IVP-Lab}

Team IVP-Lab proposes a hybrid model employing both text and image information to estimate the quality of the generated image. Initially, the image and its corresponding text are processed using the models mentioned in~\cite{radford2021learning, gu2020giqa, ko2020quality, yuan2023pku, li2023agiqa} to map image and text information into feature vectors. Then, in order to align the text and image feature vectors, multiple mapping layers are considered. In the next stage, two quality values are computed: one measuring the similarity of the image to the text (conceptual\_Q) and the other representing solely the image quality independent of the text (Image\_Q). Finally, these two values are combined to calculate the final quality score. In fact, the hybrid model assesses the quality of generated images and provides final scores using both text and image information. There are numerous factors that are important in assessing the quality of AI-generated images. These include the image’s natural appearance or naturalness, maintaining structural information, and effectively preserving the conceptual relevance to the image’s content. In the proposed approach, naturalness and structural information are represented by Image\_Q factor, and conceptual relevance is estimated using conceptual\_Q.

\subsection{Video Track}

\begin{figure*}[ht]
\centering
\includegraphics[width=\textwidth]{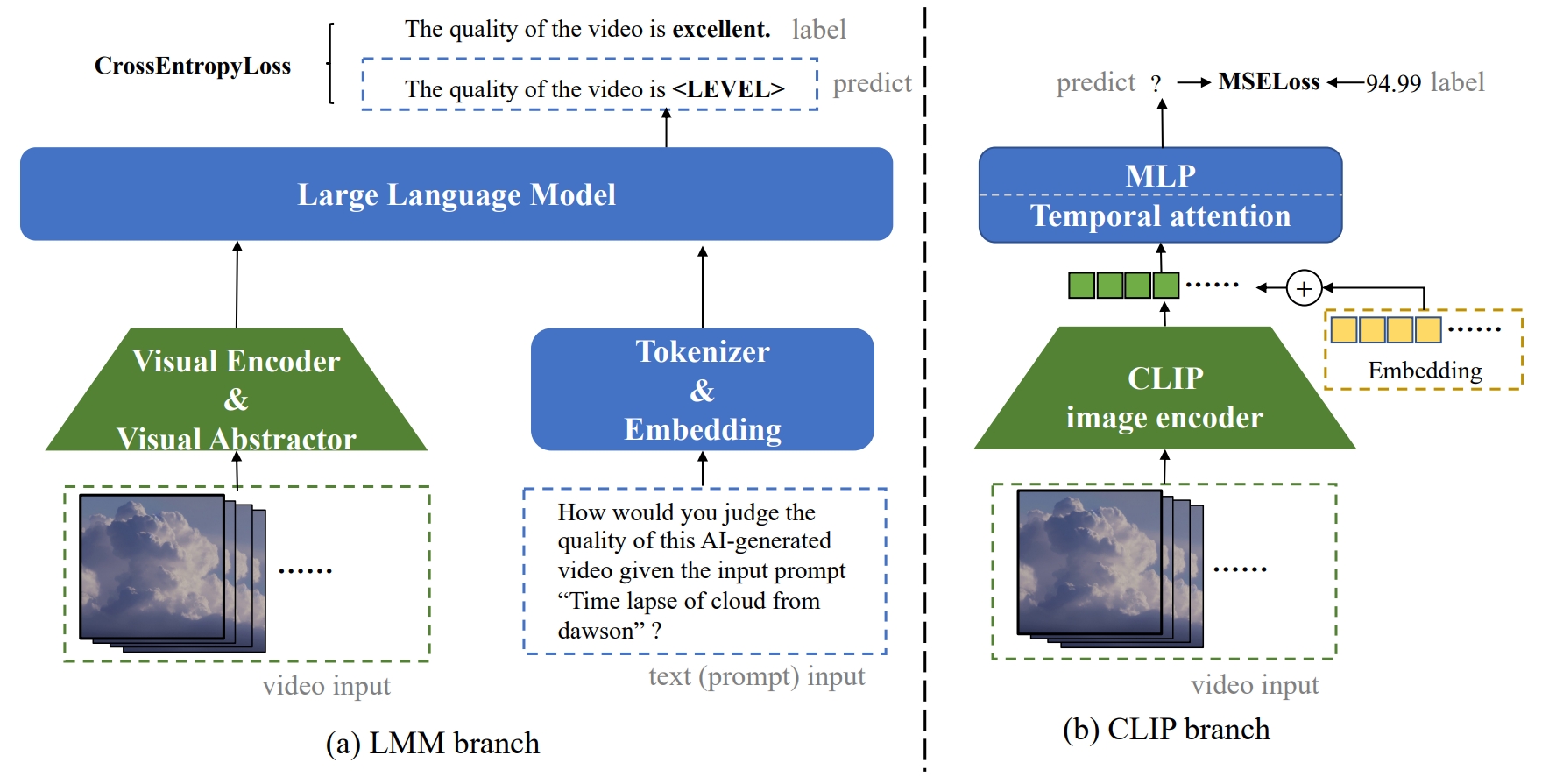}
\caption{The overview of team Kwai-kaa proposed LMM and CLIP branches.}
\label{fig:kwaikaa}
\end{figure*}

\subsubsection{IMCL-DAMO}
Team IMCL-DAMO~\cite{lu2024ntire} is the final winner of the video track. They propose a versatile video quality evaluator for AI-generated content, which can learn technical quality, aesthetic quality, 
and text-video alignment from different priors, as shown in Figure~\ref{fig:imcl}. Specifically, the input video is pre-processed to handle the disentangled information extraction from the three perspectives (i.e., technical quality, aesthetic, text-video alignment): they utilize the fragments extracted from original videos for technical quality assessment and resize the videos for aesthetic assessment and text-video alignment. Then, these separate inputs pass through multiple branches (technical branch, aesthetic branch, and alignment branch) to obtain the related score for different perspectives. To fuse the scores from different prior, we simply add them. Finally, they use  PLCC loss and rank loss for score regression of each branch.

During training, they train the technical branch and aesthetic branch by loading the pre-trained weight from LSVQ~\cite{ying2021patch}. Then the alignment branch is trained with 40\% unfixed parameters, loading the pre-trained weight from ImageReward~\cite{xu2024imagereward}. Note that these datasets are not involved in training with the provided video dataset. Finally, they finetune the technical branch, aesthetic branch, and alignment branch with 85\% unfixed parameters for late fusion. During testing, they test their network using videos provided on the official website. A self-ensemble strategy is used during testing, and it brings performance gains of about 0.008 on PLCC.

 In the training phase, the input frames for the aesthetic and text-video alignment branches are resized to $224\times224$. And ``fragments'' are sampled for the technical branch like in DOVER~\cite{wu2023dover}. They use the Swin Transformer~\cite{liu2021swin} as the technical branch backbone, the ConvNeXt~\cite{liu2022convnet} as the aesthetic branch, and BLIP~\cite{li2022blip} as the alignment branch. The training process takes 12 hours on 4 V100 GPUs. During testing, it takes 4 seconds for each video including the ensemble strategy.

\subsubsection{Kwai-kaa}
% They propose a method of evaluating the perceived quality of a video by using two separate neural networks. 
Team Kwai-kaa wins second place in the video track. They propose to tackle the challenge by leveraging LMMs. They follow a similar design to Q-Align~\cite{wu2023qalign}, which is based on mPLUG-Owl2~\cite{ye2023mplug}, with the exception of the conversation formats. Specifically, they reformulate the conversation for AI-generated video assessment as follows:

\textit{\#User:} \texttt{<video>} \textit{How would you judge the quality of this AI-generated
video given the input prompt} \texttt{<prompt>} \textit{?}

\textit{\#Assistant: The quality of the video is} \texttt{<level>}\textit{.}

In the context of conversation, \texttt{<video>} denotes the input video, \texttt{<prompt>} denotes the prompt used to generate the input video, and \texttt{<level>} denotes the predicted score by LMMs.

However, the exclusive utilization of the Q-Align~\cite{wu2023qalign} achieves limited performance due to the quantification of MOS into 5 discrete text-defined levels. This quantization strategy restricts the model’s ability to learn more precise quality scores. To complement the Q-Align architecture, they introduce an additional CLIP-based~\cite{radford2021clip} architecture, leveraging it as a robust feature extractor for predicting precise quality scores with MSE constraints. The original CLIP was tailored for images and cannot capture the temporal consistency and interconnectedness between video frames, which significantly influences video quality. To address this, they incorporate attention layers between frames to capture temporal relationships. As the prompt serves as a global abstract of the video, they believe that assessing alignment is sufficient within the LMM branch and, therefore, does not utilize text information within the CLIP branch. By employing diverse architectures and training strategies across different branches, they aim to enhance the variety of information and contribute to improved results. The final score is obtained by averaging the results of each branch. The overview of LMM and CLIP branches is shown in Figure~\ref{fig:kwaikaa}.

In particular, the LMM branches are finetuned with the pre-trained One-Align~\cite{wu2023qalign} weights, and the CLIP branches are finetuned with the pre-trained clip-vit-large-patch14 weights. They employed two finetuning strategies for the LMM branch: Vision Encoder \& Vision Abstractor (VEVA) finetuning and full model finetuning. The pre-trained model of the LMM branch is fine-tuned with an initial learning rate of $2 \times 10^{-5}$, gradually decreasing to 0 using a cosine scheduler. Each strategy is trained for 2 epochs. The precise labels are
divided into five text-defined levels: \texttt{<excellent>}, \texttt{<good>}, \texttt{<fair>}, \texttt{<poor>}, and \texttt{<bad>}. During training, they utilize 8 Tesla V100 GPUs, with a batch size of 24 for VEVA fine-tuning and 8 for full fine-tuning. The fine-tuning process for the LMM branch takes approximately 1 hour each to complete. The CLIP backbone is trained for 20 epochs, with a learning rate set to $1 \times 10^{-6}$. The training process is carried out on 8 Tesla A100 GPUs, with a batch size of 8. It takes approximately 8 hours to complete the training process. The provided generated videos consist of 4-second sequences at 4 frames per second and all frames are provided as input to the models. For videos with 15 frames, they pad the last frame to generate a complete set of 16 frames. All frames are resized to $448 \times 448$ and training processes are finished with the AdamW optimizer.

They get the MOS values of LMM branches via the weighted average of the LMM-predicted probabilities for each rating level, which can be denoted as:

\begin{equation}
    s = \mathbf{w^Tp} = \sum_{i=1}^5 w_i \times p_i = \sum_{i=1}^5 w_i \times \frac{e^{l_i}}{\sum_{j=1}^5 e^l_j},
\end{equation}

where $w_i$ is the logit weight for text level $i$ and $l_i$ is the corresponding logit output. they set the value of $w$ to $[1, 0.75, 0.5, 0.25, 0]$ for text-label \texttt{<excellent>}, \texttt{<good>}, \texttt{<fair>}, \texttt{<poor>}, and \texttt{<bad>}. They rescale the predicted scores of the CLIP branches to $[0, 1]$ by dividing them with a constant factor of 100. To get the final score, they combined the scores from different branches using a weighted approach.  Each video takes approximately 1 second to process in a single LMM branch in the inference, while the CLIP branch requires approximately 1.46 seconds per video.

\subsubsection{SQL}

\begin{figure}[t]
\centering
\includegraphics[width=\linewidth]{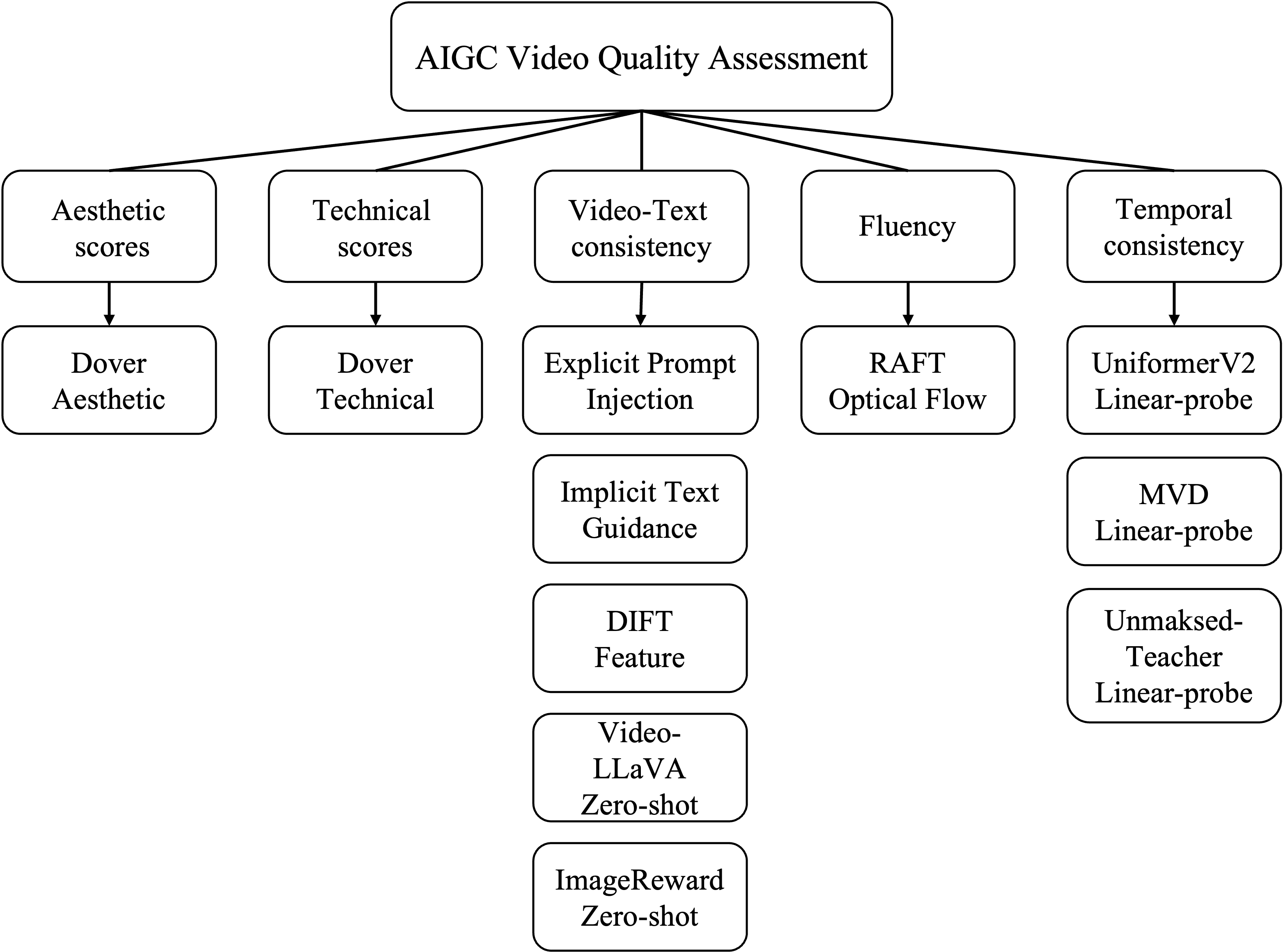}
\caption{Five dimensions for SQL team proposed method.}
\label{fig:sql}
\end{figure}

Team SQL~\cite{qu2024explore} wins third place in the video track. They propose to evaluate the video quality of AIGVs from five dimensions: aesthetic scores, technical scores, video-text consistency, fluency, and temporal consistency, as shown in Figure~\ref{fig:sql}. They refer
to aesthetic and technical aspects as visual harmony and refer to fluency and temporal consistency as temporal dynamics. Additionally, they apply model assembling and domain distribution estimation to optimize the model performance. They referred to DOVER~\cite{wu2023dover} for the aesthetic and technical evaluation of the videos. To measure video-text consistency, they apply explicit prompt injection, implicit text guidance, DIFT~\cite{tang2023dift} feature, and caption similarity. They inject the corresponding prompts of the videos into the video features using cross-attention. They also utilized BVQI’s implicit text method~\cite{wu2023exploring} and jointly optimized the evaluation network using both implicit text and explicit prompts. They further use the diffusion feature to measure the per-frame text-content consistency. Building upon this, they utilize the cross-modal text-video multi-modal large language model, Video-LLaVA~\cite{lin2023video} to generate additional captions for each video segment. They then calculate the similarity between the generated captions and the given prompts to further optimize the network. As for fluency, they incorporate an optical flow estimation module to measure the flicker degree between keyframes in the videos. After estimating the optical flow between keyframes using the RAFT~\cite{teed2020raft} algorithm, they assess video stability by computing the flow magnitude and warp loss. Regarding temporal consistency, they introduced pre-trained video understanding backbones such as Uniformer-V2~\cite{li2022uniformerv2}, UnmaskedTeacher~\cite{li2023unmasked}, and MVD~\cite{wang2023masked} to extract robust spatiotemporal features. 

In regard to model assembly, they incorporate additional five models into their main model: FAST-VQA~\cite{wu2022fast}, Faster-VQA~\cite{wu2023neighbourhood}, ZoomVQA~\cite{zhao2023quality}, XCLIP~\cite{ni2022expanding}, and ImageReward~\cite{xu2024imagereward}. It’s important to note that the integration of these models serves as a finishing touch, and they haven’t thoroughly adjusted the weights of the integration. The enhancement they bring is usually found in three decimal places. 

Considering the domain distribution differences in the results generated by different models, in supervised learning, they predict not only the score but also which text-to-video model generated the video. This additional prediction aids the model in better understanding video features. Experiments have shown that this significantly enhances the performance of their model.

\subsubsection{musicbeer}

Team musicbeer and team geniuswwg in the image track are the same teams. They use the same architecture in the video track as shown in Figure~\ref{fig:musicbeer}. Detailed information can refer to Section~\ref{musicbeer}. By changing the visual input from image to video frames, the model is able to predict the quality of AIGVs.  

% proposes an innovative approach to assess the quality of AIGC by treating it as a regression task under specified prompt conditions. It employs a dual-source CLIP~\cite{radford2021clip} text encoder to interpret prompts, combining their features for a nuanced understanding. Video features are extracted using models like ConvNeXt-Small, pre-trained on ImageNet, and adapted for interaction with the text features and vision features. A feature mixer module then blends the text and video features, using dot product and concatenation to model their correlation and conditional relationship. The final quality score is predicted by a two-layer MLP. To enhance generalization, the system applies light data augmentations like flips and brightness adjustments. The ensemble method further refines the assessment by blending predictions from three diverse models, normalized for consistency, and averaged to produce a robust quality evaluation. The method shares the same architecture as shown in Figure~\ref{fig:musicbeer}.

\subsubsection{finnbingo}

Team finnbingo and team pengfei in the image track are the same teams. They use the same architecture as introduced in Section~\ref{pengfei}.
%Team finnbingo's approach refines the LIQE~\cite{zhang2023liqe} methodology by incorporating the relationship between prompts and the images generated in AIGC tasks. To quantify this relationship, they craft specific textual templates, such as "an image how matches the prompt," where ``how'' is replaced by adverbs representing different levels of correlation: 'badly', 'poorly', 'fairly', 'well', and 'perfectly'. These templates are processed by the text encoder of the CLIP model to extract textual features, while the images are analyzed by the CLIP image encoder to obtain visual features. Leveraging CLIP adeptness at determining correlations between textual and visual features, they categorize the prompt-image alignment into five distinct levels. These correlations are then integrated using a weighted softmax function to calculate a comprehensive score. To accommodate variations in image dimensions, their method processes not only segments of images but also a resized full image to capture overarching semantics.
To adapt their methodology for video quality assessment, they treat a video as a collection of $N$ frames, from which they selectively extract $n$ frames at regular intervals to represent the video. They then calculate the average score of these $n$ frames to determine the overall quality score of the video.

%Moreover, they observed that the criteria for assessing image quality in AIGC diverge from traditional methods. While conventional evaluations typically focus on clarity, distortion, and purity, AIGC assessments are predominantly influenced by compatibility with prompts, complicating direct comparisons across different prompt-image pairs. This discrepancy renders conventional ranking losses ineffective. Instead, applying an L1 loss to directly estimate score values, thereby indirectly learning their ordinal relationship, results in improved Spearman correlation coefficients.

\subsubsection{PromptSync}

The method proposed by team PromptSync is structured into three main components. At the video level, building upon the FAST-VQA~\cite{wu2022fast} framework, they introduce additional feature constraints. They use the CLIP~\cite{radford2021clip} image encoder to obtain image features for all video frames and apply a cross-attention mechanism to detect semantic discontinuities between frames. Furthermore, they project the video features derived from FAST-VQA into the text feature space obtained from the CLIP text encoder, calculating the semantic consistency between the video and the prompt. The final AIGC evaluation score is aggregated from the FAST-VQA video features, video-prompt consistency features, and frame sequence consistency features. 

At the segment level, they replace the backbone model with Swin Transformer~\cite{liu2021swin}, concatenate text features from the CLIP text encoder, and utilize image features extracted by Swin Transformer and video features from the slowfast~\cite{feichtenhofer2019slowfast} model. The entire network is pre-trained on the LSVQ~\cite{ying2021patch} dataset and then finetuned on the competition’s training set.

Last, they conduct a frame-level evaluation of video quality. By leveraging the CLIP text encoder, they extract text features and concatenate them with image features obtained from the convent model to perform scoring. By capturing video features from different perspectives, these three components collectively contribute to their comprehensive AI-generated video quality assessment score.

\subsubsection{QA-FTE}

%Team QA-FTE proposes a vision-language fused video quality evaluator, which is designed for AIGC. Considering the quality of AI-generated video is affected by the consistency of vision and language, they use CLIP as the backbone model. Specifically, they first use the CLIP encoder to extract the vision feature of the video frame and the language feature of the prompt. Then, a bilinear pooling is used to obtain an interactive feature, which represents the consistency between vision information and language information. The above features are fused to predict AI-generated video frame quality scores. Finally, they average the scores of all frames to get an overall quality score for the video.

In the video track, team QA-FTE uses the same method as in the image track (Section~\ref{qafte}). They change the visual input from image to video frames and average the scores of all frames to get an overall quality score for AIGVs.

\subsubsection{MediaSecurity\_SYSU\&Alibaba}

Team MediaSecurity\_SYSU\&Alibaba's solution ensemble consists of four types of models: single-modal model with a single frame, single-modal model with multiple frames, multi-modal
model with a single frame, and multi-modal model with multiple frames. 

In the single-modal model with a single frame, they utilize the Swin-L~\cite{liu2021swin} pre-trained on ImageNet 22K~\cite{deng2009imagenet} to predict the quality of a single frame. In the single-modal model with multiple frames, they add NeXtVLAD~\cite{lin2018nextvlad} to the Swin-L model, which is initialized from the single-modal model with a single frame. In the multi-modal model with a single frame, they utilize multiple combinations of image encoder and text encoder, including ConvNeXt-xlarge~\cite{liu2022convnet} and Bert-base~\cite{devlin2018bert} (max length$=64$) fused by VisualBert~\cite{li2019visualbert}, Swin large and Bert-base (max length$=64$) fused by VisualBert, ConvNeXt-large and Bert-base (max length$=64$) fused by concatenation, ConvNeXt-large and DeBERTaV3-base~\cite{he2021debertav3} (max length$=64$) fused by concatenation, ConvNeXt-xlarge and Bert-base (max length$=32$) fused by concatenation, and Swin large and Bert-base (max length$=64$) fused by concatenation. Finally, in the multi-modal model with multiple frames, they use two combinations, in terms of Swin large and Bert-base (max length$=64$) fused by VisualBert,  and ConvNeXt-large and Bert-base (max length$=64$) fused by concatenation, both initialized by weights from the multi-modal model with a single frame. The final score is obtained by the ensemble of all the predictions.

\subsubsection{IPPL-VQA}

The architecture proposed by IPPL-VQA is composed of text branches and image branches. The input of the text branch is the text description of the image, and text features are extracted by the frozen text encoder of the pre-trained CLIP-B-32 model~\cite{radford2021clip}. There are two ways of sampling the image part (MaxVQA method~\cite{wu2023maxvqa}): 1. Sampling distinct texture details through cropping and splicing fragments; 2. Scaled sampling containing global information. The images of these two sampled branches undergo a frozen CLIP image encoder and two different temporal fusion models respectively. The features of both branches are concatenated and reduced to the width of the textual features with an MLP layer. The inner product of the final text and image features are calculated to get a matching score.

\subsubsection{IVP-Lab}

The proposed method of team IVP-Lab represents a hybrid model that incorporates both textual and visual data to evaluate the quality of the generated video. The mentioned model is employed to process the video and its related text, mapping the video and textual data into feature vectors. Multiple mapping layers are employed to align the text and video-based feature vectors. 

In evaluating the quality of AI-generated videos, several factors should be considered. These include the videos’s natural appearance considering both spatial and temporal information and preservation of structural information especially in the spatial domain. Another important factor is the conceptual relevance of the video’s content. 

In the proposed method, two quality-based feature vectors are computed: one assesses the similarity of the video to the text, while the other evaluates the video quality independently. These two feature vectors are then subjected to an inner product, resulting in a final vector for quality assessment. The resultant quality-based feature vector is fed to the fully connected network to estimate the quality of the AI-generated Videos.

\subsubsection{Oblivion}

Team Oblivion uses the Video Swin Transformer~\cite{liu2022video} as the visual backbone, and then they use the CLIP~\cite{radford2021clip} text encoder as the text feature extractor, using text features to enhance visual understanding to evaluate the quality of the video. They use the Swin-tiny network pre-trained on the Kinetics-400~\cite{kay2017kinetics} dataset to initialize the Video Swin Transformer backbone, and the ResNet-50~\cite{he2016deep} network pre-trained on the Kinetics-400 dataset to initialize the text encoder in the CLIP model.

%\subsubsection{CUC-IMC}

\subsubsection{UBC DSL Team}

UBC DSL Team aims to build a video quality assessment model leveraging multi-faceted video representations, taking the visual quality, text prompt, and motion coherence into account. Specifically, they use the off-the-shelf pre-trained video encoder VideoMAE~\cite{tong2022videomae} and text encoder CLIP~\cite{radford2021clip} to extract vision and language features. Additionally, they use the Inflated 3D Convnet~\cite{carreira2017quo} (I3D) as another video feature extractor, following prior work on generated video quality assessment. 

To address temporal inconsistencies in AI-generated data, such as unnatural movements or blurring, they emphasize the importance of motion coherence in video quality evaluation. Leveraging the pre-trained motion tracking network PIPs++~\cite{zheng2023pointodyssey}, they extract motion features by tracking key points’ trajectories in videos. They calculate the velocity and acceleration of these points, underpinning the notion that realistic motions should exhibit consistent acceleration. This approach yields dense motion features, enriching their VQA model’s ability to detect and interpret temporal anomalies effectively.

After extracting video and text representations using various encoders, they use a vanilla transformer consisting of an encoder layer only to mix these representations in the token space. They freeze all pre-trained encoders to prevent overfitting. They add an additional global token to improve network capacity and use the global token to read out the final
prediction score.

\section*{Acknowledgments}

We thank Huawei Technology Co., Ltd for sponsoring this NTIRE 2024 challenge and the NTIRE 2024 sponsors: ETH Z\"urich (Computer Vision Lab) and University of W\"urzburg (Computer Vision Lab).

\appendix

\section{NTIRE 2024 Organizers}
\noindent\textit{\textbf{Title: }}\\ NTIRE 2024 Quality Assessment of AI-Generated Content Challenge\\
\noindent\textit{\textbf{Members:}}\\ 
 \textit{Xiaohong Liu$^1$ (xiaohongliu@sjtu.edu.cn)}, Xiongkuo Min$^1$, Guangtao Zhai$^1$, Chunyi Li$^1$, Tengchuan Kou$^1$, Wei Sun$^1$, Haoning Wu$^2$, Yixuan Gao$^1$,  Yuqin Cao$^1$, Zicheng Zhang$^1$, Xiele Wu$^1$, Radu Timofte$^{3,4}$\\
\noindent\textit{\textbf{Affiliations: }}\\
$^{1}$ Shanghai Jiao Tong University, China\\
$^{2}$ Nanyang Technological University, Singapore\\
$^{3}$ ETH Z\"urich, Switzerland\\
$^{4}$ University of W\"urzburg, Germany

\section{Teams and Affiliations in Image Track}
\label{sec:apd:track1team}

\renewcommand{\thefootnote}{\fnsymbol{footnote}}
\subsection*{z6}
\noindent\textit{\textbf{Title:}}\\
AI-Generated Image Quality Assessment with Image and Text Feature Mixture Network\\
\noindent\textit{\textbf{Members: }}\\
Ganzorig Gankhuyag$^1$ \textit{(gnzrg25@gmail.com)}, Kihwan Yoon$^1$\\
\noindent\textit{\textbf{Affiliations: }}\\
$^1$ Korea Electronics Technology Institute\\

\renewcommand{\thefootnote}{\fnsymbol{footnote}}
\subsection*{BDVQAGroup}
\noindent\textit{\textbf{Title:}}\\
AI-Generated Image Quality Assessment Method based on LMMs\\
\noindent\textit{\textbf{Members: }}\\
Yifang Xu$^1$ \textit{(xuyifang.233@bytedance.com)}, Haotian Fan$^1$, Fangyuan Kong$^1$\\
\noindent\textit{\textbf{Affiliations: }}\\
$^1$ ByteDance\\

\renewcommand{\thefootnote}{\fnsymbol{footnote}}
\subsection*{Oblivion}
\noindent\textit{\textbf{Title:}}\\
Text-prompts to enhancement image quality assessment’s performance\\
\noindent\textit{\textbf{Members: }}\\
Shiling Zhao$^1$\textit{(yiyiaiou@163.com)}, Weifeng Dong$^1$, Haibing Yin$^1$\\
\noindent\textit{\textbf{Affiliations: }}\\
$^1$ Hangzhou Dianzi University\\

\renewcommand{\thefootnote}{\fnsymbol{footnote}}
\subsection*{MT-AIGCQA}
\noindent\textit{\textbf{Title:}}\\
AIGCQA-Bench: A Comprehensive Benchmark Suite for Text-to-image Generative Models\\
\noindent\textit{\textbf{Members: }}\\
Li Zhu$^1$ \textit{(zhuli09@meituan.com)}, Zhiling Wang$^1$, Bingchen Huang$^1$\\
\noindent\textit{\textbf{Affiliations: }}\\
$^1$ Sankuai\\

\renewcommand{\thefootnote}{\fnsymbol{footnote}}
\subsection*{pengfei}
\noindent\textit{\textbf{Title:}}\\
AIGC image quality assessment via image-prompt correspondence\\
\noindent\textit{\textbf{Members: }}\\
Fei Peng$^1$ \textit{(pf0607@bupt.edu.cn)}, Huiyuan Fu$^1$, Anlong Ming$^1$, Chuanming Wang$^1$, Huadong Ma$^1$, Shuai He$^1$, Zifei Dou$^2$, Shu Chen$^2$\\
\noindent\textit{\textbf{Affiliations: }}\\
$^1$ Beijing University of Posts and Telecommunications, China\\
$^2$ Beijing Xiaomi Mobile Software Co., Ltd.\\

\renewcommand{\thefootnote}{\fnsymbol{footnote}}
\subsection*{QA-FTE}
\noindent\textit{\textbf{Title:}}\\
Vision-Language Fused Image Quality Evaluator for AI-Generated Content\\
\noindent\textit{\textbf{Members: }}\\
Tianwu Zhi$^1$ \textit{(zhitianwu@bytedance.com)}, Yabin Zhang$^1$, Yaohui Li$^1$, Yang Li$^1$, Jingwen Xu$^1$, Jianzhao Liu$^1$, Yiting Liao$^1$, Junlin Li$^1$\\
\noindent\textit{\textbf{Affiliations: }}\\
$^1$ Bytedance Multimedia Lab\\

\renewcommand{\thefootnote}{\fnsymbol{footnote}}
\subsection*{Yag}
\noindent\textit{\textbf{Title:}}\\
AIGCIQA Implemented via Swin Transformer V2 and PickScore\\
\noindent\textit{\textbf{Members: }}\\
Zihao Yu$^1$ \textit{(yuzihao@mail.ustc.edu.cn)}, Fengbin Guan$^1$, Yiting Lu$^1$, Xin Li$^1$\\
\noindent\textit{\textbf{Affiliations: }}\\
$^1$ University of Science and Technology of China\\

\renewcommand{\thefootnote}{\fnsymbol{footnote}}
\subsection*{IVP-Lab}
\noindent\textit{\textbf{Title:}}\\
AI-generated assessment of image quality by combining textual and visual
characteristics\\
\noindent\textit{\textbf{Members: }}\\
Hossein Motamednia$^1$ \textit{(h.motamednia@ipm.ir)}, S. Farhad Hosseini-Benvidi$^2$, Ahmad Mahmoudi-Aznaveh$^3$ and Azadeh Mansouri$^2$\\
\noindent\textit{\textbf{Affiliations: }}\\
$^1$ High Performance Computing Laboratory, School of Computer Science, Institute for Research in Fundamental Sciences, Tehran, Iran \\
$^2$ Department of Electrical and Computer Engineering, Faculty of Engineering, Kharazmi University, Tehran, Iran\\
$^3$ Cyber Research Institute, Shahid Beheshti University, Tehran, Iran\\

\renewcommand{\thefootnote}{\fnsymbol{footnote}}
\subsection*{IQ Analyzers}
\noindent\textit{\textbf{Title:}}\\
Quality Assessment of AI-Generated Content Using Bag of Features Approach\\
\noindent\textit{\textbf{Members: }}\\
Avinab Saha$^1$ \textit{(avinab.saha@utexas.edu)}, Sandeep Mishra$^1$, Shashank Gupta$^1$, Rajesh Sureddi$^1$, Oindrila Saha$^2$\\
\noindent\textit{\textbf{Affiliations: }}\\
$^1$ University of Texas at Austin\\
$^2$ University of Massachusetts Amher\\

\renewcommand{\thefootnote}{\fnsymbol{footnote}}
\subsection*{IVL}
\noindent\textit{\textbf{Title:}}\\
Quality Assessment of AI-Generated Contents through Language-Image Pre-trained Models and Support Vector Regression\\
\noindent\textit{\textbf{Members: }}\\
Luigi Celona$^1$ \textit{(luigi.celona@unimib.it)}, Simone Bianco$^1$, Paolo Napoletano$^1$, Raimondo Schettini$^1$\\
\noindent\textit{\textbf{Affiliations: }}\\
$^1$ Department of Informatics Systems and Communication, University of Milano - Bicocca\\

\renewcommand{\thefootnote}{\fnsymbol{footnote}}
\subsection*{HUTB-IQALab}
\noindent\textit{\textbf{Title:}}\\
Mixture-of-Experts Boosted Visual Perception and Semantic-Aware Quality Assessment for AI-Generated Images\\
\noindent\textit{\textbf{Members: }}\\
Junfeng Yang$^1$ \textit{(b12100031@hnu.edu.cn)}, Jing Fu$^1$, Wei Zhang$^1$, Wenzhi Cao$^1$, Limei Liu$^1$, Han Peng$^1$\\
\noindent\textit{\textbf{Affiliations: }}\\
$^1$ Xiangjiang Laboratory and Hunan University of Technology and Business\\

\renewcommand{\thefootnote}{\fnsymbol{footnote}}
\subsection*{JNU\_620}
\noindent\textit{\textbf{Title:}}\\
Prompt-StairIQA\\
\noindent\textit{\textbf{Members: }}\\
Weijun Yuan$^1$ \textit{(yweijun@stu2022.jnu.edu.cn)}, Zhan Li$^1$, Yihang Cheng$^1$, Yifan Deng$^1$\\
\noindent\textit{\textbf{Affiliations: }}\\
$^1$ Jinan University\\

\renewcommand{\thefootnote}{\fnsymbol{footnote}}
\subsection*{MediaSecurity\_SYSU\&Alibaba}
\noindent\textit{\textbf{Title:}}\\
Single modal and multiple multi-modal networks for learning quality assessment\\
\noindent\textit{\textbf{Members: }}\\
Huacong Zhang$^1$ \textit{(zhanghc8@mail2.sysu.edu.cn)}, Haiyi Xie$^1$, Chengwei Wang$^1$, Baoying Chen$^2$, Jishen Zeng$^2$, Jianquan Yang$^1$\\
\noindent\textit{\textbf{Affiliations: }}\\
$^1$ Sun Yat-sen University\\
$^2$ Alibaba Group\\

\renewcommand{\thefootnote}{\fnsymbol{footnote}}
\subsection*{geniuswwg}
\noindent\textit{\textbf{Title:}}\\
PCQA: A Strong Baseline for AIGC Quality Assessment Based on Prompt Condition\\
\noindent\textit{\textbf{Members: }}\\
Weigang Wang$^1$ \textit{(geniuswwg@gmail.com)}, Xi Fang$^2$, Xiaoxin Lv$^3$, Jun Yan$^4$\\
\noindent\textit{\textbf{Affiliations: }}\\
$^1$ Cisco Systems, Inc. \\
$^2$ DP Technology, Ltd. \\
$^3$ Shopee Pte. Ltd. \\
$^4$ Tongji University

\renewcommand{\thefootnote}{\fnsymbol{footnote}}
\subsection*{PKUMMCAL}
\noindent\textit{\textbf{Title:}}\\
Assessing AI-Generated Image Quality via Multitask Learning\\
\noindent\textit{\textbf{Members: }}\\
Haohui Li$^1$ \textit{(lihaohui@stu.pku.edu.cn)}, Bowen Qu$^1$, Yao Li$^1$, Shuqing Luo$^1$, Shunzhou Wang$^1$, Wei Gao$^1$\\
\noindent\textit{\textbf{Affiliations: }}\\
$^1$ School of Electronic and Computer Engineering, Shenzhen Graduate School, Peking University\\

\renewcommand{\thefootnote}{\fnsymbol{footnote}}
\subsection*{CVLab}
\noindent\textit{\textbf{Title:}}\\
An Efficient CLIP-based Baseline for Evaluating the Alignment and Quality of AI-Generated Images\\
\noindent\textit{\textbf{Members: }}\\
Zihao Lu$^1$ \textit{(zihao.lu@stud-mail.uni-wuerzburg.de)}, Marcos V. Conde$^1$, Radu Timofte$^1$\\
\noindent\textit{\textbf{Affiliations: }}\\
$^1$ University of W{\"u}rzburg

\section{Teams and Affiliations in Video Track}

\renewcommand{\thefootnote}{\fnsymbol{footnote}}
\subsection*{IMCL-DAMO}
\noindent\textit{\textbf{Title:}}\\
VC-VQE: Versatile and Comprehensive Video Quality Evaluator for AI-Generated Content\\
\noindent\textit{\textbf{Members: }}\\
Yiting Lu$^1$ \textit{(luyt31415@mail.ustc.edu.cn)}, Xin Li$^1$, Xinrui Wang$^1$, Zihao Yu$^1$, Fengbin Guan$^1$, Zhibo Chen$^1$, Ruling Liao$^1$, Yan Ye$^1$\\
\noindent\textit{\textbf{Affiliations: }}\\
$^1$ University of Science and Technology of China\\

\renewcommand{\thefootnote}{\fnsymbol{footnote}}
\subsection*{Kwai-kaa}
\noindent\textit{\textbf{Title:}}\\
Leveraging Large Multi-modality Models and CLIP Encoder for AI-Generated Video Assessment\\
\noindent\textit{\textbf{Members: }}\\
 Qiulin Wang$^1$ \textit{(wangqiulin@kuaishou.com)}, Bing Li$^2$, Zhaokun Zhou$^3$, Miao Geng$^1$, Rui Chen$^1$, Xin Tao$^1$\\
\noindent\textit{\textbf{Affiliations: }}\\
$^1$ Kuaishou Technology\\
$^2$ University of Science and Technology of China\\
$^3$ Peking University\\

\renewcommand{\thefootnote}{\fnsymbol{footnote}}
\subsection*{SQL}
\noindent\textit{\textbf{Title:}}\\
Exploring AIGC Video Quality: A Focus on Text-Video Consistency, Visual Harmony, and Temporal Dynamics\\
\noindent\textit{\textbf{Members: }}\\
 Wei Gao$^1$ \textit{(gaowei262@pku.edu.cn)}, Xiaoyu Liang$^1$, Bowen Qu$^1$, Shangkun Sun$^1$\\
\noindent\textit{\textbf{Affiliations: }}\\
$^1$ Peking University\\

\renewcommand{\thefootnote}{\fnsymbol{footnote}}
\subsection*{musicbeer}
\noindent\textit{\textbf{Title:}}\\
PCQA: A Strong Baseline for AIGC Quality Assessment Based on Prompt Condition\\
\noindent\textit{\textbf{Members: }}\\
Xiaoxin Lv$^1$ \textit{(musicbeer2017@gmail.com)}, Xi Fang$^2$, Weigang Wang$^3$, Jun Yan$^4$\\
\noindent\textit{\textbf{Affiliations: }}\\
$^1$ Shopee Pte. Ltd. \\
$^2$ DP Technology, Ltd. \\
$^3$ Cisco Systems, Inc. \\
$^4$ Tongji University

\renewcommand{\thefootnote}{\fnsymbol{footnote}}
\subsection*{finnbingo}
\noindent\textit{\textbf{Title:}}\\
Video Quality Assessment for AI-Generated Content via Frame-Prompt Correspondence\\
\noindent\textit{\textbf{Members: }}\\
  Xingyuan Ma$^1$ \textit{(maxy@bupt.edu.cn)}, Shuai He$^1$, Anlong Ming$^1$, Huiyuan Fu$^1$, Huadong Ma$^1$, Zifei Dou$^2$, Shu Chen$^2$\\
\noindent\textit{\textbf{Affiliations: }}\\
$^1$ Beijing University of Posts and Telecommunications\\
$^2$ Beijing Xiaomi Mobile Software Co., Ltd\\

\renewcommand{\thefootnote}{\fnsymbol{footnote}}
\subsection*{PromptSync}
\noindent\textit{\textbf{Title:}}\\
multi-modal AIGC Video Quality Assessment with CLIP and Swin Transformer\\
\noindent\textit{\textbf{Members: }}\\
 Jiaze Li$^1$ \textit{(1916444377@qq.com)}, Mengduo Yang$^1$, Haoran Xu$^1$, Jie Zhou$^1$, Shiding Zhu$^1$, Bohan Yu$^1$\\
\noindent\textit{\textbf{Affiliations: }}\\
$^1$ Zhejiang University\\

\renewcommand{\thefootnote}{\fnsymbol{footnote}}
\subsection*{QA-FTE}
\noindent\textit{\textbf{Title:}}\\
Vision-Language Fused Video Quality Evaluator for AI-Generated Content\\
\noindent\textit{\textbf{Members: }}\\
Tianwu Zhi$^1$ \textit{(hitianwu@bytedance.com)}, Yabin Zhang$^1$, Yaohui Li$^1$, Yang Li$^1$, Jingwen Xu$^1$, Jianzhao Liu$^1$, Yiting Liao$^1$, Junlin Li$^1$\\
\noindent\textit{\textbf{Affiliations: }}\\
$^1$ Bytedance Inc.\\

\renewcommand{\thefootnote}{\fnsymbol{footnote}}
\subsection*{MediaSecurity\_SYSU\&Alibaba}
\noindent\textit{\textbf{Title:}}\\
Integrating 2D and 3D Modalities for Enhanced multi-modal Video Quality Assessment\\
\noindent\textit{\textbf{Members: }}\\
Baoying Chen$^1$ \textit{(chenbaoying.chenba@alibaba-inc.com)}, Jishen Zeng$^1$, Huacong Zhang$^1$,  Haiyi Xie$^2$, Chengwei Wang$^2$,  Jianquan Yang$^2$\\
\noindent\textit{\textbf{Affiliations: }}\\
$^1$ Alibaba Group\\
$^2$ Sun Yat-sen University\\

\renewcommand{\thefootnote}{\fnsymbol{footnote}}
\subsection*{IPPL-VQA}
\noindent\textit{\textbf{Title:}}\\
XCS-Net\\
\noindent\textit{\textbf{Members: }}\\
Pengfei Chen$^1$ \textit{(chenpengfei@xidian.edu.cn)}, Xinrui Xu$^1$, Jiabin Shen$^1$, Zhichao Duan$^1$\\
\noindent\textit{\textbf{Affiliations: }}\\
$^1$ Xidian University\\

\renewcommand{\thefootnote}{\fnsymbol{footnote}}
\subsection*{IVP-Lab}
\noindent\textit{\textbf{Title:}}\\
AI-generated assessment of video quality by combining textual and visual characteristics\\
\noindent\textit{\textbf{Members: }}\\
Hossein Motamednia$^1$ \textit{(h.motamednia@ipm.ir)}, S. Farhad Hosseini-Benvidi$^2$, Erfan Asadi$^2$, Ahmad Mahmoudi-Aznaveh$^3$, Azadeh Mansouri$^2$\\
\noindent\textit{\textbf{Affiliations: }}\\
$^1$ Institute for Research in Fundamental Sciences\\
$^2$ Kharazmi University\\
$^3$ Shahid Beheshti University\\

\renewcommand{\thefootnote}{\fnsymbol{footnote}}
\subsection*{Oblivion}
\noindent\textit{\textbf{Title:}}\\
Text-prompts to enhancement video quality assessment’s performance\\
\noindent\textit{\textbf{Members: }}\\
Weifeng Dong$^1$\textit{(dongwf@hdu.edu.cn)}, Shiling Zhao$^1$, Haibing Yin$^1$\\
\noindent\textit{\textbf{Affiliations: }}\\
$^1$ Hangzhou Dianzi University\\

% \renewcommand{\thefootnote}{\fnsymbol{footnote}}
% \subsection*{CUC-IMC}
% \noindent\textit{\textbf{Title:}}\\
% \\
% \noindent\textit{\textbf{Members: }}\\
% Zelu Qi, Da Pan, Zefeng Ying, Shuqi Wang, Xiaofeng Liu, Fei Zhao\\
% \noindent\textit{\textbf{Affiliations: }}\\
% \\

\renewcommand{\thefootnote}{\fnsymbol{footnote}}
\subsection*{UBC DSL Team}
\noindent\textit{\textbf{Title:}}\\
Enhanced Video Quality Assessment Transformer Based on Motion Feature\\
\noindent\textit{\textbf{Members: }}\\
Jiahe Liu$^1$\textit{(jiaheliu@ece.ubc.ca)}, Qi Yan$^1$, Youran Qu$^2$, Xiaohui Zeng$^3$, Lele Wang$^1$, Renjie Liao$^1$\\
\noindent\textit{\textbf{Affiliations: }}\\
$^1$ University of British Columbia\\
$^2$ Peking University\\
$^3$ University of Toronto

%%%%%%%%% REFERENCES

{\small
\bibliographystyle{ieee_fullname}
\bibliography{egbib}
}

\end{document}